\documentclass{article} 
\usepackage{iclr2027_conference,times}

\usepackage{amsmath,amsfonts,bm}

\def\eqref#1{equation~\ref{#1}}

\def\1{\bm{1}}

\DeclareMathAlphabet{\mathsfit}{\encodingdefault}{\sfdefault}{m}{sl}
\SetMathAlphabet{\mathsfit}{bold}{\encodingdefault}{\sfdefault}{bx}{n}

\usepackage{hyperref}
\usepackage{url}

\usepackage[utf8]{inputenc}
\usepackage[T1]{fontenc}
\usepackage{graphicx}
\usepackage{tikz}
\usetikzlibrary{arrows.meta,backgrounds,calc,fit,positioning}
\usepackage{amsthm}
\usepackage{mathtools}
\usepackage{algorithm}
\usepackage{algpseudocode}
\usepackage{booktabs}
\usepackage{colortbl}
\usepackage{comment} 
\usepackage{hyperref}
\usepackage{url}
\usepackage{xcolor}
\usepackage{enumitem}
\usepackage{soul}
\usepackage{capt-of}
\usepackage{subcaption}

\usepackage{multirow}
\usepackage[table]{xcolor}

\usepackage{adjustbox}

\definecolor{bestaccent}{HTML}{2F5D8C}
\definecolor{scopeaccent}{HTML}{18794E}
\definecolor{blockgray}{HTML}{EEEEEE}

\title{SCOPD: Sparse-Context On-Policy Self-Distillation for Efficient Vision-Language Models}

\author{%
\bf Ahmadreza Jeddi$^{1,2*}$ \quad Enming Zhang$^{1,2*}$ \quad Jasper Gerigk$^{1,2}$ \quad Hakki Karaimer$^{3}$ \\
\bf Mozhgan Nasr Azadani$^{4,5}$ \quad Jiayun Luo$^{2,6}$ \quad Minh Ngoc Le$^{1}$ \quad Gholamali Aminian$^{7}$ \\
\bf Hugo Buurmeijer$^{4}$ \quad Yongchao Chen$^{8}$ \quad Leonid Sigal$^{2,6}$ \quad Igor Gilitschenski$^{1,2}$ \\
\bf Konstantinos G. Derpanis$^{2,9}$ \quad Marco Pavone$^{4,10}$ \quad Babak Taati$^{1,2}$ \\[6pt]
$^{1}$University of Toronto \quad $^{2}$Vector Institute \quad $^{3}$Samsung AI Center Toronto \\
$^{4}$Stanford University \quad $^{5}$University of Waterloo \quad $^{6}$University of British Columbia \\
$^{7}$Alan Turing Institute \quad $^{8}$Tsinghua University \quad $^{9}$York University \quad $^{10}$NVIDIA \\[4pt]
$^{*}$Equal contribution
}

\iclrfinalcopy 

\begin{document}

\maketitle
\lhead{Preprint. Under review.}


 \definecolor{customred}{HTML}{ED028C}  

\begin{abstract}
Reasoning vision-language models (VLMs) process images and videos as long sequences of visual tokens, making inference expensive. Training-free token pruning can substantially reduce this cost, but performance degrades sharply under aggressive compression, commonly attributed to irreversible loss of task-relevant visual information. We show that this explanation is incomplete. In a fixed-context Pass@$K$ analysis, we prune each visual representation once and repeatedly sample reasoning trajectories from the same sparse context. Although greedy Pass@1 drops substantially, Pass@$K$ recovers many otherwise failed examples, suggesting that useful visual evidence can remain accessible but is not reliably utilized during reasoning. We call this the \emph{representation--utilization gap}. Motivated by this observation, we introduce \textsc{SCOPD}, a sparse-context on-policy self-distillation framework in which a student model generates reasoning trajectories from pruned visual tokens while a privileged full-context teacher supervises those on-policy prefixes. \textsc{SCOPD} requires no ground-truth responses, architectural modifications, or inference-time computation. Building on \textsc{SCOPD}, we develop \textsc{SCOPD+} to focus distillation where visual evidence matters most. A small visual-budget intervention identifies positions most sensitive to additional visual evidence and selectively distills them while backpropagating through only a fraction of response positions. At $10\%$ visual-token retention, the Vanilla base model retains only $86.37\%$ of the unpruned model's performance across 13 benchmarks. \textsc{SCOPD} raises this normalized aggregate to $90.49\%$, while \textsc{SCOPD+} further improves it to $92.43\%$. Across token budgets, benchmarks, and pruning operators, our results show that efficient reasoning VLMs depend not only on \emph{which} visual information survives pruning, but also on how reliably the language model learns to \emph{use} the sparse representation that remains. Project Page: \href{https://armenjeddi.github.io/scopd/}{\textcolor{customred}{https://armenjeddi.github.io/scopd/}}
\end{abstract}


\captionsetup[subfigure]{
    font=normalsize,
    labelfont=normalfont,
    textfont=normalfont,
    justification=centering,
    singlelinecheck=false,
    skip=3pt
}

\begin{figure*}[t]
    \centering

    \begin{minipage}[t]{0.42\linewidth}
        \vspace{0pt}
        \centering

        \begin{subfigure}[t]{\linewidth}
            \centering

            \includegraphics[
                width=\linewidth
            ]{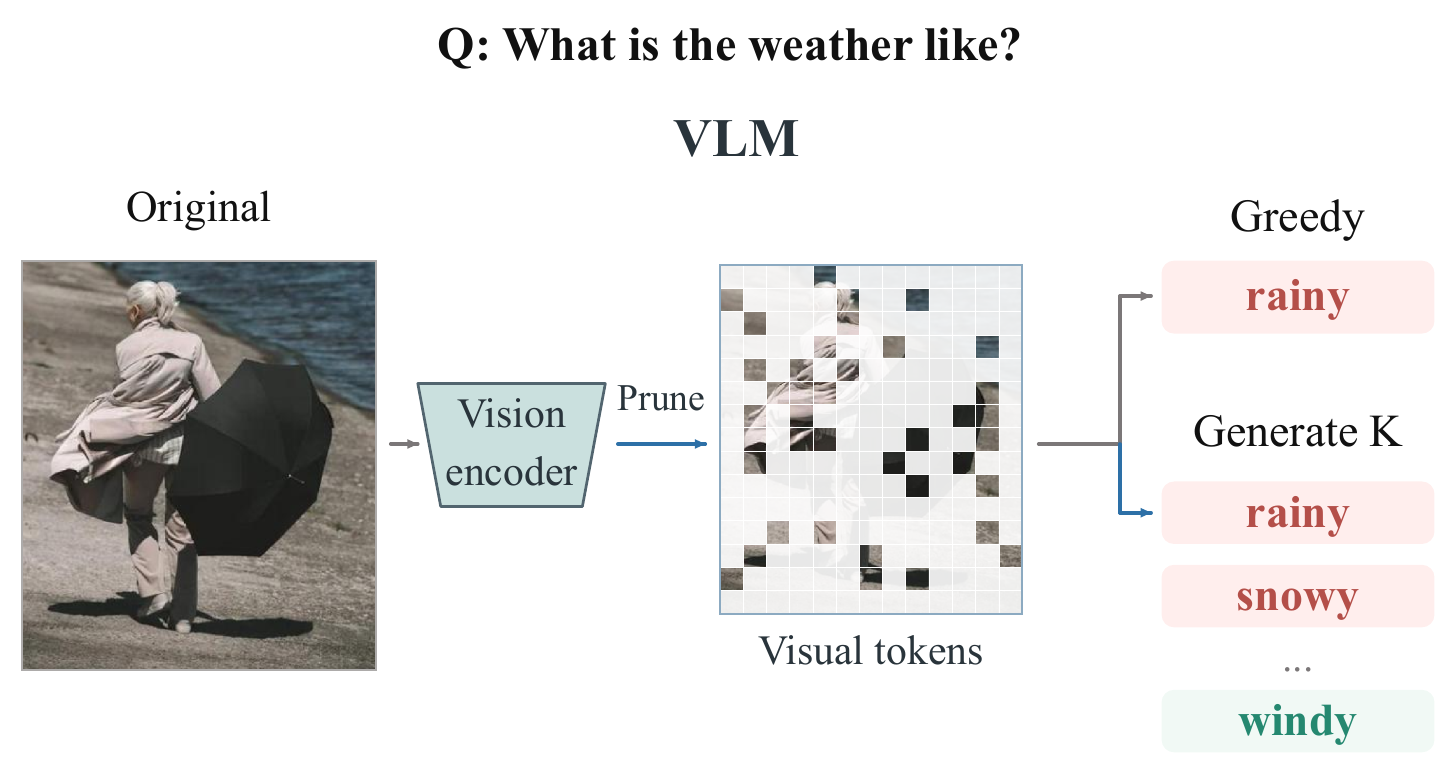}

            \vspace{-0.25em}

            \caption{
                Visual evidence is still here
            }
            \label{fig:recoverability-a}
        \end{subfigure}

        \vspace{0.7em}

        \begin{subfigure}[t]{\linewidth}
            \centering

            \includegraphics[
                width=\linewidth
            ]{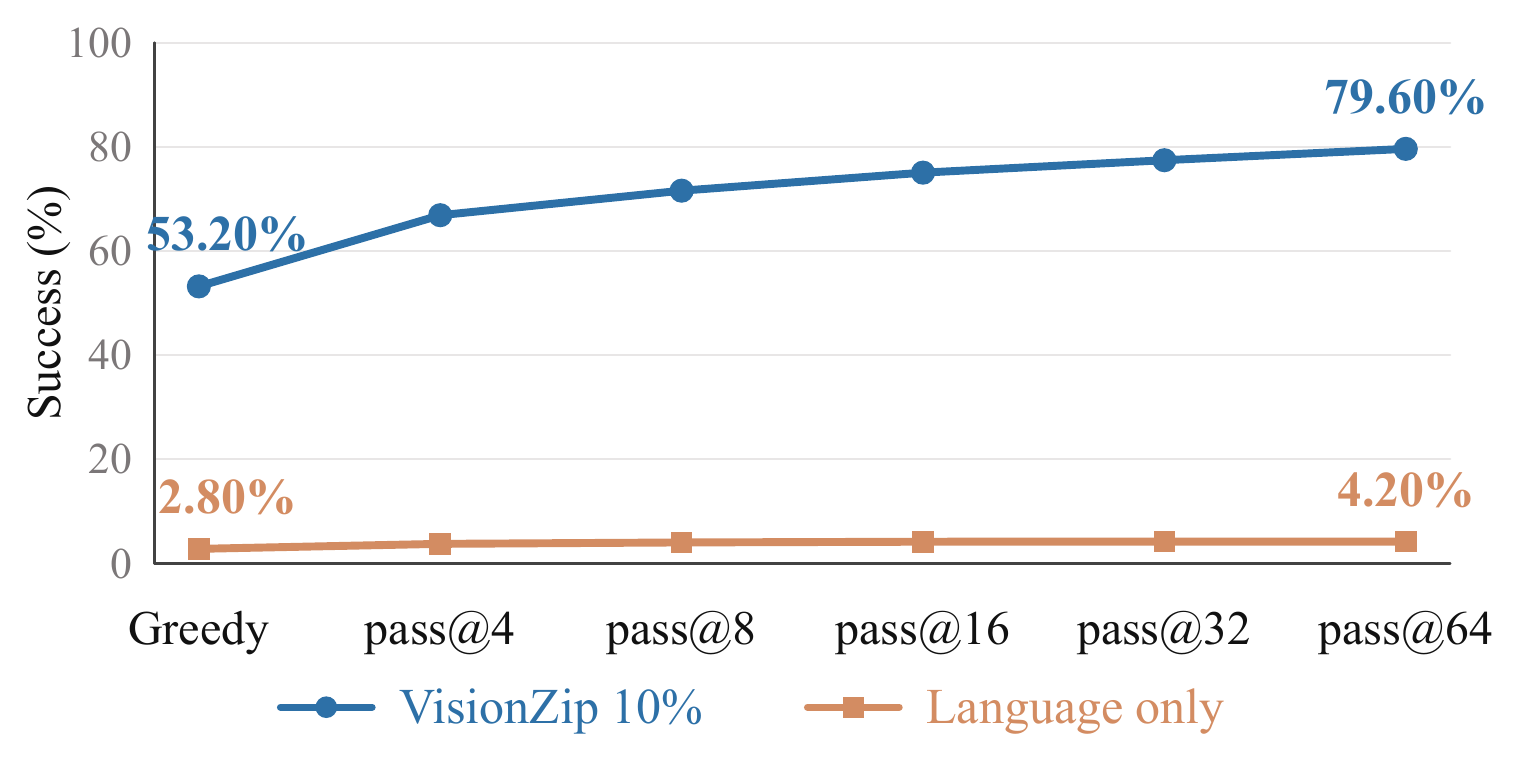}

            \vspace{0.15em}

            \caption{
                Pass@$K$ reveals utilization gap
            }
            \label{fig:recoverability-b}
        \end{subfigure}

    \end{minipage}
    \hfill
    \begin{subfigure}[t]{0.56\linewidth}
        \vspace{0pt}
        \centering

        \includegraphics[
            width=0.96\linewidth
        ]{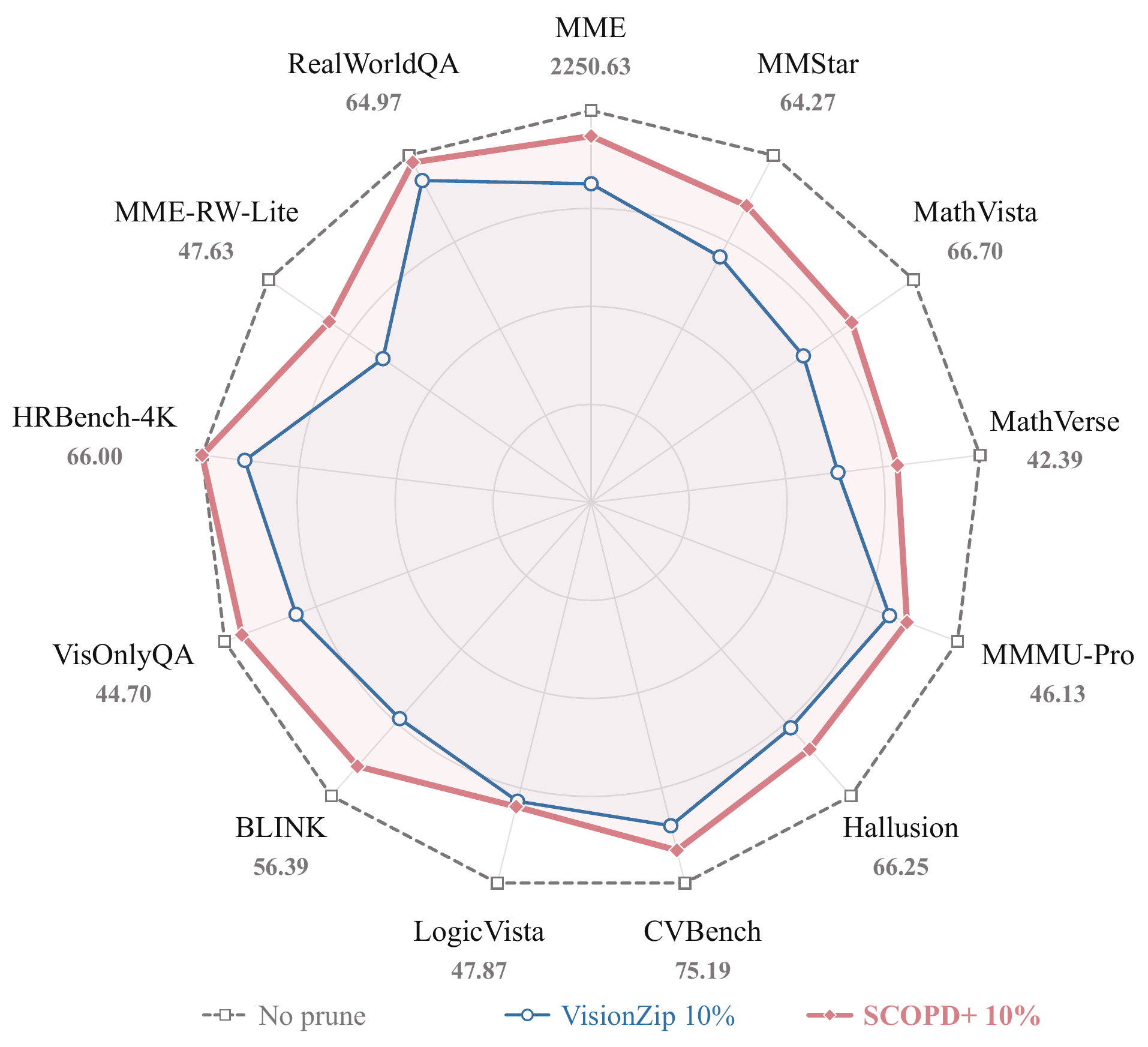}

        \vspace{0.15em}

        \caption{
            \textsc{SCOPD+} recovers the utilization gap
        }
        \label{fig:recoverability-c}

    \end{subfigure}


    \caption{
\textbf{Visual-token pruning creates a representation--utilization gap that sparse-context adaptation can recover.}
\textbf{(a)} Repeated sampling from the same sparse visual context can recover valid reasoning trajectories missed by greedy decoding.
\textbf{(b)} Across 500 examples solved greedily by the unpruned model, Pass@$K$ substantially improves under aggressive VisionZip~\cite{yang2025visionzip} pruning, while a language-only control remains near zero, showing that useful visual evidence remains accessible.
\textbf{(c)} \textsc{SCOPD+} complements token pruning by adapting the model to better use this remaining evidence, substantially improving performance at aggressive token budgets.
}

    \vspace{-0.5em}

    \label{fig:recoverability-concept}

\end{figure*}

\section{Introduction}

Reasoning vision-language models (VLMs) encode images and videos into hundreds or thousands of visual tokens, making language-model prefill a major source of inference cost. A growing body of work therefore seeks to reduce visual context through token pruning, merging, or compression~\cite{chen2024image,li2023blip,shang2025llava}. While these methods can remove many visual tokens with modest degradation, performance often drops sharply under aggressive compression. This drop is commonly attributed to the loss of task-relevant visual information during pruning.

To test whether pruning failures necessarily imply information loss, we perform a \emph{fixed-context Pass@$K$}~\cite{chen2021evaluating} analysis on 500 image--question pairs. For each example, we prune the visual representation once and then draw multiple stochastic reasoning trajectories while keeping the image, question, model, and retained visual tokens unchanged; full experimental details and analysis are provided in Sec.\ \ref{sec:representation_utilization}. As illustrated in Figs.~\ref{fig:recoverability-a} and~\ref{fig:recoverability-b}, aggressive pruning causes a large drop in greedy Pass@1, yet repeated sampling recovers many correct solutions from the same sparse representation. Since the retained visual tokens are fixed across all generations, this recovery cannot be explained by obtaining a more favorable pruned context. Instead, it suggests that task-relevant evidence can remain accessible, but the language model fails to consistently access it during reasoning. We refer to this discrepancy as the \emph{representation--utilization gap}.

This observation motivates adapting the language model to its deployment-time sparse visual context. We introduce \textsc{SCOPD} (\textbf{S}parse-\textbf{C}ontext \textbf{O}n-\textbf{P}olicy self-\textbf{D}istillation), an on-policy self-distillation formulation tailored to visual-token pruning. The student generates reasoning trajectories using the sparse representation, while a privileged teacher with access to the corresponding full visual context supervises the same student-generated prefixes. This provides dense on-policy supervision from the privileged visual context. We find that this straightforward asymmetric use of sparse and full visual context recovers a substantial fraction of the performance lost under aggressive pruning, with no architectural changes or additional inference-time overhead.

Yet dense supervision does not necessarily correspond to \emph{visual} supervision. Large teacher--student disagreement can occur at reasoning positions only weakly influenced by visual context, making KL divergence alone a poor indicator of where privileged visual supervision is most useful. We therefore introduce a \emph{visual sensitivity} signal measuring how much the student's next-token distribution changes under a small increase in visual evidence. To focus \textsc{SCOPD}'s supervision where visual evidence matters most, we develop \textsc{SCOPD+}, a selective variant that distills visually sensitive reasoning positions while backpropagating through only a fraction of response tokens.

Across 13 image benchmarks at $10\%$ visual-token retention, the Vanilla model retains $86.37\%$ of its unpruned performance, which increases to $90.49\%$ with \textsc{SCOPD} and $92.43\%$ with \textsc{SCOPD+}. These gains generalize across pruning operators based on attention, token diversity, and even random selection, as well as across Qwen2.5-VL and Qwen3-VL and from image training to video evaluation without video-specific post-training. Neither method requires supervised reasoning traces, architectural modifications, or additional inference-time computation. Together, these results show that learning to better utilize the visual evidence that survives pruning can substantially boost the performance of reasoning VLMs under sparse visual context.

Our contributions are:
\begin{itemize}[leftmargin=*,labelindent=0pt,label=\textbf{--}]
    \item We identify a \emph{representation--utilization} gap in
vision-language models under visual-token pruning: models can retain sufficient visual information in sparse representations yet fail to effectively use that information during reasoning.

    \item We introduce \textsc{SCOPD}, a sparse-context on-policy self-distillation framework in which the student generates reasoning trajectories from pruned visual tokens, while a full-context teacher supervises the same on-policy prefixes, without requiring ground-truth responses or reasoning traces.

    \item We extend \textsc{SCOPD} with \textsc{SCOPD+}, a selective variant that uses a small visual-budget intervention to identify visually sensitive response tokens, achieving the strongest performance while backpropagating through only a fraction of response positions.
\end{itemize}

\section{Related Work}
\label{sec:related_work}

\paragraph{Efficient Visual Contexts for VLMs.}
Modern VLMs often include learned interfaces that already compress visual features before they reach the language model, including resampling, projection, and compact token representations~\cite{alayrac2022flamingo,li2023blip,li2025tokenpacker}. Nevertheless, high-resolution images and especially long videos can still produce large visual contexts, motivating more aggressive visual-token reduction~\cite{li2024llama,vasu2025fastvlm,zhang2025llava}. A large body of recent work performs post-hoc pruning or merging using attention, similarity, diversity, graph structure, or language-conditioned relevance, often without modifying or retraining the underlying VLM~\cite{chen2024image,shang2025llava,zhang2024sparsevlm,jeddi2025similarity,xing2024pyramiddrop,jeddi2026avis,yang2025visionzip,alvar2025divprune}. Other approaches introduce learned or layer-adaptive sparsification modules, trading greater flexibility for additional training or architectural changes~\cite{huang2025dynamic,ye2025atp,zeng2026glimpse,wang2026hawaii,ivanovic2025efficient}. EPIC further studies the \emph{training difficulty induced by visual-token compression}, using progressive token- and layer-level consistency distillation to help the model adapt to the compressed feature space~\cite{wen2026efficient}. These works primarily study \emph{which} tokens should be retained or how models should adapt to compressed representations. A recent analysis also shows that aggregate benchmark scores can hide important pruning failures on vision-centric tasks~\cite{endo2025feather}. Our work instead asks whether degradation after pruning necessarily reflects missing visual evidence, or whether surviving evidence remains available but is used unreliably.

\paragraph{On-Policy and Privileged Self-Distillation.}
Classical knowledge distillation matches teacher and student predictions on a fixed data distribution~\cite{hinton2015distilling,kim2016sequence}, but autoregressive models can suffer from state-distribution mismatch because generation depends on the student's own previous outputs. Generalized Knowledge Distillation addresses this by querying the teacher on student-generated trajectories~\cite{agarwal2024policy}, while on-policy RL methods such as GRPO optimize samples from the current policy using sparse outcome-level rewards~\cite{shao2024deepseekmath,guo2025deepseek}. Recent On-Policy Self-Distillation (OPSD) instead provides dense token-level supervision: the student generates on-policy, while a stronger conditional view of the same model acts as teacher~\cite{zhao2026self,shenfeld2026self}. This principle has recently been extended to multimodal reasoning using privileged crops, resolutions, or fine-grained visual evidence~\cite{yuan2026vision,zhu2026rp,liu2026visual,tian2026vicur,venkatraman2026perception,li2026visual}. Our setting differs in that the model is unchanged while its conditioning representation is deliberately sparsified. \textsc{SCOPD} uses the corresponding unpruned representation as privileged supervision to recover reasoning ability that remains latent under the sparse context.

\paragraph{Selective and Efficient On-Policy Distillation.}
Recent work observes that dense teacher supervision is not equally useful across all tokens or trajectories. Methods such as TIP~\cite{xu2026tip} identify informative positions using uncertainty and teacher--student disagreement, while others restrict distillation when teacher guidance becomes unreliable on drifted prefixes~\cite{fu2026revisiting}. In VLMs, Visual-Advantage OPD reweights supervision to emphasize visually grounded reasoning, while trajectory-level approaches select or repair entire reasoning paths rather than uniformly matching every token~\cite{liu2026visual,jiang2026trajectory}. \textsc{SCOPD+} differs in how it identifies useful privileged supervision: rather than relying on teacher--student disagreement alone, it directly perturbs the student's available visual evidence while holding the reasoning prefix fixed. The resulting distributional change isolates positions sensitive to missing visual context, separating visually relevant disagreement from ordinary language-level variation.

\paragraph{Recoverability Beyond Greedy Decoding.}
Pass@$K$ and self-consistency have long shown that greedy decoding can underestimate capabilities revealed by repeated sampling~\cite{chen2021evaluating,wang2022self}. Inference-time scaling work similarly studies how successful-response coverage grows with larger sampling budgets~\cite{brown2024large,snell2024scaling,chen2026does}. Most closely related, ShortOPD shows that structurally pruned LLMs can suffer a large Pass@1 drop while retaining substantial Pass@$K$, suggesting useful generations may be demoted rather than erased~\cite{zhang2026shortopd,wen2026benchmark}. We instead study pruning of the conditioning signal rather than the model itself. By holding both the model and sparse visual representation fixed across generations, we isolate whether failures arise because visual evidence was removed or because surviving evidence is used unreliably, directly motivating our representation--utilization gap.

\section{Adapting VLMs to Sparse Visual Contexts}
\label{sec:method}

We investigate whether degradation under visual-token pruning necessarily reflects the loss of task-relevant information~(Sec.\ \ref{sec:representation_utilization}). Fixed-context repeated sampling shows that correct reasoning trajectories often remain recoverable despite sharp drops in greedy performance, revealing a \emph{representation--utilization gap}. To adapt the language model to sparse visual contexts and address this gap, we introduce \textsc{SCOPD}~(Sec.\ \ref{sec:SCOPD}) for sparse-context on-policy self-distillation and extend it with \textsc{SCOPD+}~(Sec.\ \ref{sec:SCOPD-plus}) for selective supervision of visually sensitive response tokens.

\subsection{Motivation: The Representation--Utilization Gap}
\label{sec:representation_utilization}

Performance loss under aggressive visual-token pruning is commonly attributed to removing task-relevant evidence. While this certainly occurs, performance may also degrade because sufficient evidence remains in
the sparse representation but is no longer used reliably by the language model during
autoregressive decoding. We therefore ask: \emph{when a reasoning VLM fails after pruning, can a correct reasoning trajectory still be recovered from the exact same sparse visual context?}

\paragraph{Fixed-context recoverability analysis.}
We construct 500 image--question pairs from MMStar~\cite{chen2024we}, CVBench~\cite{tong2024cambrian}, MMMU-Pro~\cite{yue2025mmmu}, BLINK~\cite{fu2024blink}, LogicVista~\cite{xiao2024logicvista}, and LLaVA-CoT~\cite{xu2025llava}, restricted to free-form numerical questions that the unpruned model solves correctly under greedy decoding. For each example, we prune the visual representation once with VisionZip~\cite{yang2025visionzip}, retaining $10\%$ of tokens, and keep the resulting sparse context fixed across generations. We use VisionZip as a representative strong pruning method, while later ablations show that our adaptation generalizes across pruning operators. We then compare greedy decoding with repeated stochastic sampling up to Pass@$64$. To distinguish recovery from visual evidence from success driven by language priors, we repeat the experiment in a language-only setting with all visual input removed. Finally, to ensure that recovered answers reflect coherent, visually grounded reasoning rather than accidental answer matching, we use GPT-6-Astra-Max~\cite{openai2026gpt6astra} to verify that the reasoning is logically consistent and supported by the visual evidence. Full dataset construction, prompts, decoding settings, metrics, and additional analyses are provided in Appendix~\ref{app:pass-at-k}.

Fig.~\ref{fig:recoverability-a} illustrates the setup, while Fig.~\ref{fig:recoverability-b} quantifies the effect. At $10\%$ retention, greedy success falls to $53.2\%$, yet Pass@$64$ rises to $79.6\%$ from the \emph{same} sparse representation. By contrast, the language-only setting remains near zero, increasing from $2.8\%$ to $4.2\%$. Because the retained visual tokens are fixed across generations, this recovery cannot be attributed to a more favorable pruning outcome. Instead, the sparse representation still supports correct reasoning trajectories that the model fails to produce reliably. We refer to this mismatch as the \emph{representation--utilization gap}.

This distinction motivates adapting the language model itself to sparse visual contexts: when relevant evidence survives pruning, performance can be recovered by learning to use that representation more reliably rather than changing the pruner.

\begin{figure*}[t]
    \centering

    \begin{subfigure}[t]{0.42\linewidth}
        \vspace{0pt}
        \centering
        \includegraphics[width=\linewidth]{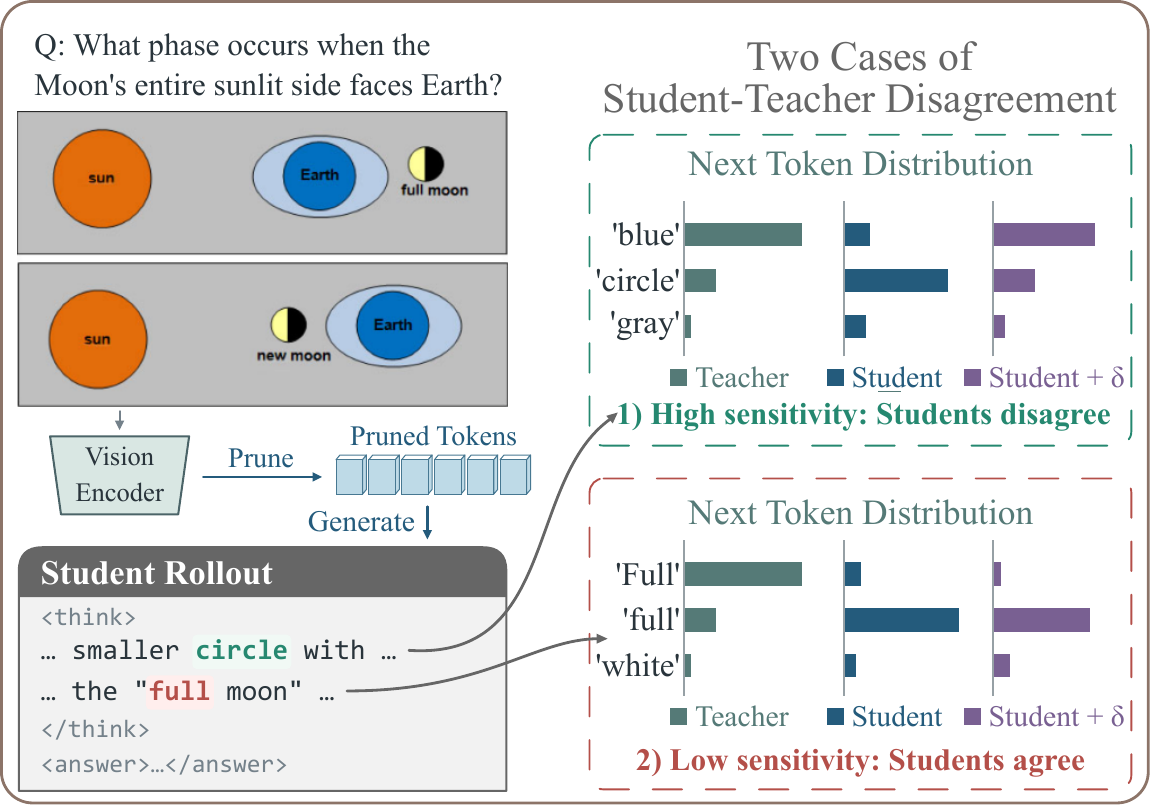}
        \caption{High Teacher--student KL alone does not reveal visual sensitivity}
        \label{fig:SCOPD-kl}
    \end{subfigure}
    \hfill
    \begin{subfigure}[t]{0.57\linewidth}
        \vspace{0pt}
        \centering
        \includegraphics[width=\linewidth]{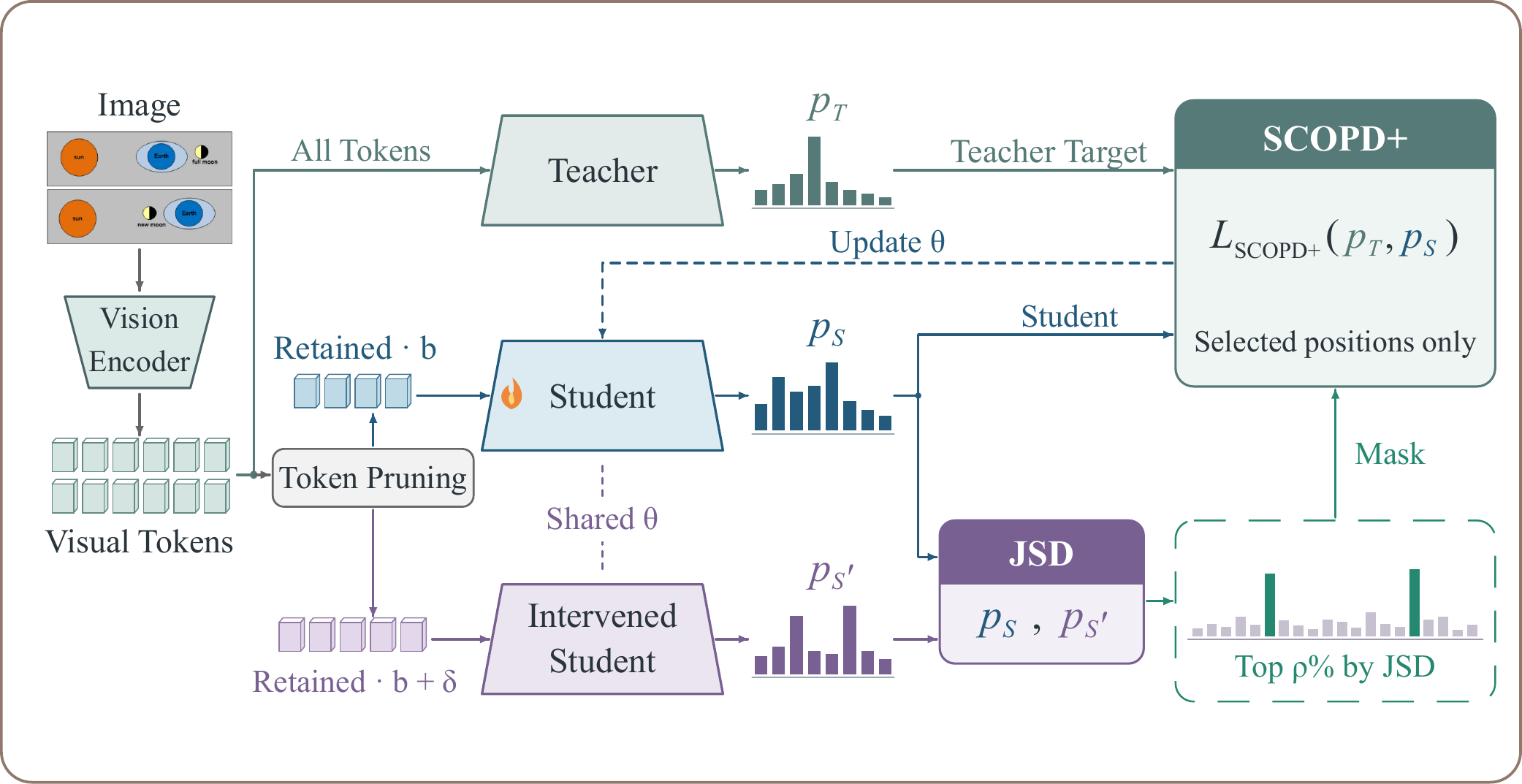}
        \caption{\textsc{SCOPD+} pipeline}
        \label{fig:SCOPD-pipeline}
    \end{subfigure}

    \caption{
    \textbf{Motivation and pipeline of \textsc{SCOPD+}.}
    \textbf{(a)} Large teacher--student KL can arise from language-level disagreement even when a prediction is weakly affected by visual context. We therefore use a small visual-budget intervention as a direct test of visual dependence: positions whose predictions change with additional visual evidence are more visually sensitive.
    \textbf{(b)} The student generates an on-policy trajectory at budget $b$. The same prefixes are scored by the student at budgets $b$ and $b^{+}$ and by the full-context teacher. We compute visual sensitivity using the Jensen--Shannon divergence between the two student distributions, select the top $\rho=10\%$ response positions, and apply the teacher-to-student KL loss only at those positions.
    }
    \label{fig:SCOPD}
\end{figure*}

\subsection{SCOPD: Sparse-Context On-Policy Self-Distillation}
\label{sec:SCOPD}

The representation--utilization gap identified in Sec.~\ref{sec:representation_utilization} suggests that, when task-relevant evidence survives pruning, the language model should be adapted to reason more reliably from its deployment-time sparse visual context. We therefore introduce \textsc{SCOPD} (\textbf{S}parse-\textbf{C}ontext \textbf{O}n-\textbf{P}olicy self-\textbf{D}istillation), an on-policy self-distillation~\cite{zhao2026self} formulation tailored to visual-token pruning. We sample reasoning trajectories under the sparse visual context, then score the same prefixes under the corresponding full visual representation to obtain token-level supervision. This aligns training with the states visited by the pruned model without requiring ground-truth answers or reasoning traces.

Let $x_i$ denote the textual input, $Z_i$ the full visual-token sequence, and $Z_i^{b}=P_b(Z_i)$ the output of pruning operator $P$ at retention budget $b$. The student samples a reasoning trajectory
\begin{equation}
    \hat{y}_i \sim p_\theta(\cdot \mid x_i,Z_i^{b}),
\end{equation}
and, at each decoding position $t$, the sparse-context student and full-context teacher score the same prefix $\hat{y}_{i,<t}$:
\begin{align}
    p^{b}_{i,t}
    &= p_\theta(\cdot \mid x_i,Z_i^{b},\hat{y}_{i,<t}), \\
    q_{i,t}
    &= p_{\bar{\theta}}(\cdot \mid x_i,Z_i,\hat{y}_{i,<t}),
\end{align}
where $\bar{\theta}$ denotes the teacher parameters. \textsc{SCOPD} minimizes
\begin{equation}
    \mathcal{L}^{\textsc{SCOPD}}_i
    =
    \frac{1}{T_i}
    \sum_{t=1}^{T_i}
    D_{\mathrm{KL}}\!\left(
        q_{i,t}\,\|\,p^{b}_{i,t}
    \right).
\end{equation}

\textsc{SCOPD} applies dense supervision at every reasoning position, but not all teacher--student disagreement is equally informative about the visual context. As illustrated in Fig.~\ref{fig:SCOPD}a, many positions with large KL divergence are only weakly affected by missing visual evidence and instead reflect ordinary language-modeling disagreement. This motivates selecting supervision based not only on teacher--student disagreement, but also on whether a prediction is actually sensitive to the available visual context.

\subsection{SCOPD+: Selective Distillation of Visually Sensitive Tokens}
\label{sec:SCOPD-plus}

\textsc{SCOPD+} extends \textsc{SCOPD} by concentrating privileged supervision on reasoning positions most sensitive to visual evidence. Dense distillation may otherwise emphasize supervision to positions with substantial teacher--student disagreement that is only weakly related to visual information. Our intuition is simple: if adding a small amount of visual information substantially changes the student's next-token distribution, that position is more likely to depend on visual context rather than primarily on language-modeling dynamics. Fig.~\ref{fig:SCOPD}b summarizes the pipeline.

For a student operating at token retention budget $b$, we construct a slightly richer visual context
\begin{equation}
    b^{+}=b+\delta,
    \qquad
    Z_i^{b^{+}}=P_{b^{+}}(Z_i).
\end{equation}
We then evaluate
\begin{equation}
    p^{b^{+}}_{i,t}
    =
    p_\theta(\cdot\mid x_i,Z_i^{b^{+}},\hat{y}_{i,<t}).
\end{equation}
The intervention introduces no new rollout: $p^{b^{+}}_{i,t}$ is evaluated on the same student-generated reasoning prefix, and uses the same encoded visual features.
The intervention therefore requires only one additional sparse visual context and language-model forward pass.

We quantify token-level visual sensitivity using Jensen--Shannon divergence (JSD)~\cite{lin1991divergence}:
\begin{equation}
\begin{aligned}
S_{i,t}
&=
\frac{1}{2}D_{\mathrm{KL}}\!\left(
p^{b}_{i,t}\middle\|m_{i,t}\right)
+
\frac{1}{2}D_{\mathrm{KL}}\!\left(
p^{b^{+}}_{i,t}\middle\|m_{i,t}\right),
\qquad m_{i,t}
=
\frac{1}{2}\left(
p^{b}_{i,t}+p^{b^{+}}_{i,t}
\right).
\end{aligned}
\end{equation}

We use JSD because the intervention compares two predictive distributions under different visual-evidence budgets without privileging either one. Unlike directional KL, JSD is symmetric and bounded, making the sensitivity score easier to compare across tokens and less dominated by extreme probability ratios. A symmetrized forward--reverse KL would also remove directionality, but remains unbounded and can be overly sensitive to low-probability events.

Large $S_{i,t}$ indicates that position $t$ is highly sensitive to a small increase in visual context, making it a promising target for adaptation. We therefore retain the top $\rho$ fraction of positions, 
\begin{equation}
    \mathcal{K}_i
    =
    \operatorname{TopK}_{t\in\{1,\ldots,T_i\}}
    \left(
        S_{i,t},
        \left\lceil\rho T_i\right\rceil
    \right),
\end{equation}
and apply full-context supervision only to these tokens:
\begin{equation}
    \mathcal{L}_{\textsc{SCOPD+},i}
    =
    \frac{1}{|\mathcal{K}_i|}
    \sum_{t\in\mathcal{K}_i}
    D_{\mathrm{KL}}\!\left(
        q_{i,t}\,\|\,p^{b}_{i,t}
    \right).
\end{equation}

Thus, \textsc{SCOPD+} preserves the same on-policy reasoning trajectories as \textsc{SCOPD}, while restricting backpropagation to positions most responsive to visual evidence. This suppresses visually uninformative teacher--student disagreement and substantially reduces the number of response tokens used for distillation.

\section{Experiments}
\label{sec:experiments}

\definecolor{tableheader}{HTML}{24364B}
\definecolor{tableline}{HTML}{A8B4C2}
\definecolor{tablebase}{HTML}{F1F4F7}
\definecolor{tableours}{HTML}{E7F5EC}
\definecolor{tableaccent}{HTML}{18794E}
\definecolor{ratiochip}{HTML}{DDE7F2}

\newcommand{\best}[1]{\textcolor{black}{\textbf{#1}}}

\subsection{Experimental Setup}
\label{sec:experimental-setup}

Unless otherwise stated, we use Qwen2.5-VL-7B-Instruct~\cite{bai2025qwen25vltechnicalreport} with VisionZip~\cite{yang2025visionzip} as the pruning operator $P$ and a default visual-token retention budget of $b=10\%$. All post-training methods use LLM-only LoRA ($r=16$, $\alpha=32$, zero dropout), a constant learning rate of $2\times10^{-5}$, and an effective batch size of 32 across four GPUs; the vision encoder and multimodal projector remain frozen. For \textsc{SCOPD} and \textsc{SCOPD+}, the teacher is an exponential moving average (EMA) of the student parameters with decay $0.9999$. For \textsc{SCOPD+}, we set $\delta=1\%$ and $\rho=10\%$ by default. We train on approximately 10K examples from LLaVA-CoT~\cite{xu2025llava}. Reference reasoning traces and answers are used only by supervised baselines, while \textsc{SCOPD} and \textsc{SCOPD+} obtain supervision online from the full-context teacher. We evaluate across a broad suite of image and video benchmarks spanning perception, hallucination, spatial understanding, mathematical and logical reasoning, and temporal video understanding. Full training, data, benchmark, and evaluation details are provided in Appendix~\ref{app:eval-train-setup}.

\paragraph{Baselines.}
We compare against: (i) \textbf{Vanilla}, the base Qwen2.5-VL model evaluated without post-training; (ii) \textbf{SFT}, supervised fine-tuning on sparse visual context using reference reasoning traces and answers; (iii) \textbf{EPIC}~\cite{wen2026efficient}, a compression-aware baseline based on progressive consistency distillation, using its original retention schedule and objective; (iv) \textbf{GRPO}, an RLVR-style on-policy baseline that generates reasoning trajectories from the same pruned visual context and optimizes them using outcome-based rewards (see Appendix~\ref{app:grpo-details}); (v) \textbf{\textsc{SCOPD}}, our sparse-context on-policy self-distillation formulation using full-context privileged supervision; and (vi) \textbf{\textsc{SCOPD+}}, which selectively applies \textsc{SCOPD} supervision to visually sensitive reasoning positions.


\definecolor{blockgray}{HTML}{F1F4F7}
\definecolor{scopeblue}{HTML}{DDEEFF}

\newcommand{\second}[1]{\underline{#1}}
\newcommand{\benchhead}[1]{{\normalsize\textbf{#1}}}
\newcommand{\score}[1]{{\normalsize #1}}

\begin{table*}[t]
\centering
\caption{
\textbf{Main results across visual-token budgets.}
We compare post-training methods on 13 image benchmarks using
Qwen2.5-VL-7B-Instruct with VisionZip at $100\%$, $20\%$, and $10\%$
visual-token retention.
Avg$_{13}$ is the mean performance normalized to the unpruned Vanilla
model on each benchmark.
Best and second-best results are \textbf{bold} and \underline{underlined}, respectively.
Overall, \textsc{SCOPD} improves sparse-context performance, with \textsc{SCOPD+} performing best under aggressive pruning.
}
\vspace{-0.4em}
\label{tab:main-results}

\setlength{\tabcolsep}{2.4pt}
\renewcommand{\arraystretch}{1.20}

\begin{adjustbox}{max width=\textwidth}
\begin{tabular}{l|ccccccccccccc|c}
\toprule

\benchhead{Method}
& \benchhead{MME}
& \benchhead{MMStar}
& \benchhead{MathVista}
& \benchhead{MathVerse}
& \benchhead{MMMU-Pro}
& \benchhead{Hallusion}
& \benchhead{CVBench}
& \benchhead{LogicVista}
& \benchhead{BLINK}
& \benchhead{VisOnlyQA}
& \benchhead{HR4K}
& \benchhead{MME-RW-Lite}
& \benchhead{RealWorldQA}
& \benchhead{Avg$_{13}$} \\
\midrule

\rowcolor{blockgray}
\multicolumn{15}{c}{
    \small\textit{Full Visual Context (100\% Tokens)}
} \\

{\small Vanilla}
& \score{2250.63}
& \score{64.27}
& \score{66.70}
& \score{\second{42.39}}
& \score{46.13}
& \score{66.25}
& \score{75.19}
& \score{47.87}
& \score{56.39}
& \score{\second{44.70}}
& \score{66.00}
& \score{47.63}
& \score{64.97}
& \score{100.00} \\

{\small SFT}
& \score{\best{2380.54}}
& \score{64.93}
& \score{65.30}
& \score{35.41}
& \score{\second{46.82}}
& \score{64.35}
& \score{\second{75.73}}
& \score{43.18}
& \score{55.18}
& \score{42.43}
& \score{\best{69.62}}
& \score{\best{52.01}}
& \score{\best{67.58}}
& \score{99.17} \\

{\small EPIC}
& \score{2315.82}
& \score{\best{65.40}}
& \score{65.60}
& \score{36.42}
& \score{\best{48.09}}
& \score{65.09}
& \score{\best{77.00}}
& \score{\best{49.22}}
& \score{54.87}
& \score{42.87}
& \score{\second{68.62}}
& \score{\second{50.96}}
& \score{66.93}
& \score{100.30} \\

{\small GRPO}
& \score{2262.94}
& \score{64.47}
& \score{67.10}
& \score{\second{42.39}}
& \score{46.18}
& \score{66.25}
& \score{75.23}
& \score{\second{48.55}}
& \score{\second{56.65}}
& \score{43.91}
& \score{67.62}
& \score{49.30}
& \score{\second{67.06}}
& \score{\best{100.84}} \\

{\small SCOPD \textbf{(Ours)}}
& \score{\second{2322.58}}
& \score{64.20}
& \score{\best{68.00}}
& \score{\best{43.02}}
& \score{45.49}
& \score{\second{68.14}}
& \score{73.20}
& \score{47.20}
& \score{\best{56.71}}
& \score{\best{45.65}}
& \score{64.62}
& \score{47.68}
& \score{\best{67.58}}
& \score{\second{100.67}} \\

\rowcolor{scopeblue}
{\small\textbf{SCOPD+ (Ours)}}
& \score{2310.91}
& \score{\second{65.20}}
& \score{\second{67.90}}
& \score{40.61}
& \score{44.34}
& \score{\best{68.98}}
& \score{73.68}
& \score{46.76}
& \score{56.02}
& \score{44.52}
& \score{64.88}
& \score{48.67}
& \score{\second{67.06}}
& \score{100.02} \\

\midrule
\rowcolor{blockgray}
\multicolumn{15}{c}{
    \small\textit{Retain 20\% Visual Tokens}
} \\

{\small Vanilla}
& \score{2106.54}
& \score{60.27}
& \score{61.60}
& \score{37.56}
& \score{43.82}
& \score{61.51}
& \score{73.12}
& \score{40.72}
& \score{52.03}
& \score{42.35}
& \score{66.50}
& \score{43.77}
& \score{62.61}
& \score{93.42} \\

{\small SFT}
& \score{\best{2368.60}}
& \score{60.53}
& \score{60.90}
& \score{32.99}
& \score{44.62}
& \score{62.57}
& \score{73.59}
& \score{38.26}
& \score{53.18}
& \score{\best{42.78}}
& \score{\best{70.38}}
& \score{\best{47.21}}
& \score{\best{67.97}}
& \score{95.22} \\

{\small EPIC}
& \score{2225.28}
& \score{59.67}
& \score{58.90}
& \score{33.63}
& \score{\best{45.55}}
& \score{60.78}
& \score{\best{76.45}}
& \score{41.16}
& \score{52.55}
& \score{42.52}
& \score{\second{69.12}}
& \score{\second{46.80}}
& \score{\second{66.54}}
& \score{94.71} \\

{\small GRPO}
& \score{2150.62}
& \score{60.07}
& \score{62.60}
& \score{37.06}
& \score{44.05}
& \score{62.88}
& \score{\second{73.73}}
& \score{40.04}
& \score{52.34}
& \score{42.00}
& \score{67.25}
& \score{44.24}
& \score{63.27}
& \score{93.95} \\

{\small SCOPD \textbf{(Ours)}}
& \score{\second{2270.21}}
& \score{\best{62.87}}
& \score{\best{65.00}}
& \score{\second{39.47}}
& \score{\second{44.97}}
& \score{\second{63.83}}
& \score{73.51}
& \score{\second{41.83}}
& \score{\second{54.60}}
& \score{\second{42.70}}
& \score{65.88}
& \score{46.01}
& \score{65.88}
& \score{\second{96.80}} \\

\rowcolor{scopeblue}
{\small\textbf{SCOPD+ (Ours)}}
& \score{2235.58}
& \score{\second{61.80}}
& \score{\second{64.50}}
& \score{\best{42.01}}
& \score{43.99}
& \score{\best{65.72}}
& \score{72.68}
& \score{\best{43.18}}
& \score{\best{55.13}}
& \score{42.35}
& \score{66.25}
& \score{46.33}
& \score{65.88}
& \score{\best{97.25}} \\

\midrule
\rowcolor{blockgray}
\multicolumn{15}{c}{
    \small\textit{Retain 10\% Visual Tokens}
} \\

{\small Vanilla}
& \score{2040.13}
& \score{54.87}
& \score{55.30}
& \score{26.90}
& \score{41.85}
& \score{58.57}
& \score{69.53}
& \score{\second{37.58}}
& \score{48.97}
& \score{40.35}
& \score{62.38}
& \score{39.19}
& \score{62.61}
& \score{86.37} \\

{\small SFT}
& \score{\best{2277.41}}
& \score{55.67}
& \score{54.90}
& \score{25.63}
& \score{40.75}
& \score{56.78}
& \score{\best{72.86}}
& \score{34.68}
& \score{50.45}
& \score{40.96}
& \score{\best{68.00}}
& \score{\best{44.66}}
& \score{\best{66.80}}
& \score{88.82} \\

{\small EPIC}
& \score{2149.19}
& \score{55.87}
& \score{50.40}
& \score{24.37}
& \score{41.45}
& \score{55.73}
& \score{\second{72.69}}
& \score{32.66}
& \score{50.13}
& \score{40.26}
& \score{65.75}
& \score{\second{43.67}}
& \score{\second{64.44}}
& \score{86.45} \\

{\small GRPO}
& \score{2061.41}
& \score{55.47}
& \score{54.90}
& \score{25.38}
& \score{\second{41.91}}
& \score{58.99}
& \score{70.08}
& \score{34.00}
& \score{48.71}
& \score{40.09}
& \score{64.25}
& \score{39.40}
& \score{61.18}
& \score{85.73} \\

{\small SCOPD \textbf{(Ours)}}
& \score{\second{2201.28}}
& \score{\second{58.60}}
& \score{\best{60.60}}
& \score{\second{30.46}}
& \score{\best{42.95}}
& \score{\second{60.46}}
& \score{72.65}
& \score{34.68}
& \score{\second{52.81}}
& \score{\second{42.09}}
& \score{63.75}
& \score{41.84}
& \score{64.31}
& \score{\second{90.49}} \\

\rowcolor{scopeblue}
{\small\textbf{SCOPD+ (Ours)}}
& \score{2177.06}
& \score{\best{59.60}}
& \score{\second{60.30}}
& \score{\best{33.38}}
& \score{\best{42.95}}
& \score{\best{60.99}}
& \score{71.96}
& \score{\best{38.26}}
& \score{\best{53.55}}
& \score{\best{43.65}}
& \score{\second{66.00}}
& \score{43.15}
& \score{64.31}
& \score{\best{92.43}} \\

\bottomrule
\end{tabular}
\end{adjustbox}

\vspace{-0.5em}

\end{table*}

\subsection{Main Results}
\label{sec:main-results}

We evaluate sparse-context post-training across 13 image benchmarks and multiple visual-token budgets. Table~\ref{tab:main-results} reports the baseline and proposed-method results. Avg$_{13}$ is the mean benchmark performance normalized to the unpruned Vanilla model.

\vspace{-0.3em}

\paragraph{Findings.}
Table~\ref{tab:main-results} highlights three trends:
\begin{itemize}[leftmargin=*,labelindent=0pt,label=\textbf{--}]
    \item \textbf{\textsc{SCOPD} effectively adapts reasoning VLMs to sparse visual context.}
    At $20\%$ and $10\%$ retention, \textsc{SCOPD} improves Avg$_{13}$ over the Vanilla baseline from $93.42\%$ to $96.80\%$ and from $86.37\%$ to $90.49\%$, respectively, outperforming SFT, GRPO, and EPIC. This shows that full-context on-policy supervision recovers a substantial fraction of pruning-induced degradation.

    \item \textbf{Selective visual supervision provides further gains.}
    \textsc{SCOPD+} reaches $97.25\%$ Avg$_{13}$ at $20\%$ retention and $92.43\%$ at $10\%$, improving over dense \textsc{SCOPD} by $+0.45$ and $+1.94$ points. At the more aggressive $10\%$ budget, \textsc{SCOPD+} improves over \textsc{SCOPD} on 8 of 13 benchmarks and matches it on two others, with particularly large normalized gains on LogicVista, MathVerse, VisOnlyQA, and HR4K. These benchmarks place strong demands on visually grounded reasoning, diagram understanding, or fine-grained visual perception, making the pattern consistent with \textsc{SCOPD+}'s emphasis on visually sensitive reasoning positions.

    \item \textbf{The gains target sparse-context adaptation rather than generic post-training improvements.}
    With the full visual context, \textsc{SCOPD} and \textsc{SCOPD+} achieve $100.67\%$ and $100.02\%$ Avg$_{13}$, respectively, remaining essentially unchanged from the unpruned Vanilla model. Their advantage instead grows as the visual budget shrinks, supporting the representation--utilization hypothesis that post-training primarily helps the model use compressed visual evidence more reliably. We further evaluate an extreme $5\%$ retention setting in Appendix~\ref{app:aggressive-pruning}, where the same trend persists.

\end{itemize}

Overall, \textsc{SCOPD} provides strong adaptation to sparse visual representations, while \textsc{SCOPD+} further improves performance by focusing supervision on visually sensitive reasoning tokens.

\vspace{-0.5em}

\subsection{Ablations}
\label{sec:ablations}

\vspace{-0.5em}

\definecolor{scopeblue}{HTML}{DDEEFF}

\begin{table*}[t]
\centering
\noindent


\begin{minipage}[t]{0.485\textwidth}
\vspace{0pt}
\centering

\captionof{table}{
\textbf{Token-selection ablation.}
Selective methods retain $\rho=10\%$ of response tokens; dense \textsc{SCOPD} uses all positions. \textsc{SCOPD+} performs best overall.}
\label{tab:ablate-sensitivity-metric}

\vspace{-0.5em}

\setlength{\tabcolsep}{1.8pt}
\renewcommand{\arraystretch}{1.20}

\begin{adjustbox}{max width=\linewidth}
\begin{tabular}{l|rrrrrr|r}
\toprule
\benchhead{Selection}
& \benchhead{MME}
& \benchhead{MMStar}
& \benchhead{BLINK}
& \benchhead{VisOnly}
& \benchhead{MathVista}
& \benchhead{RWQA}
& \benchhead{Avg$_6$} \\
\midrule

{\small Dense \textsc{SCOPD}}
& \score{\textbf{2201.28}}
& \score{58.60}
& \score{52.81}
& \score{42.09}
& \score{\textbf{60.60}}
& \score{\textbf{64.31}}
& \score{94.44} \\

\midrule

{\small Random}
& \score{2141.89}
& \score{59.87}
& \score{52.97}
& \score{41.74}
& \score{58.50}
& \score{63.66}
& \score{93.56} \\

{\small Top KL}
& \score{2174.15}
& \score{\underline{60.13}}
& \score{53.29}
& \score{42.00}
& \score{60.20}
& \score{\underline{63.79}}
& \score{\underline{94.51}} \\

{\small TIP}
& \score{\underline{2179.75}}
& \score{\textbf{60.40}}
& \score{\textbf{53.76}}
& \score{\underline{42.26}}
& \score{59.20}
& \score{62.22}
& \score{94.21} \\

{\small Bottom Sens.}
& \score{2035.59}
& \score{56.27}
& \score{48.29}
& \score{40.96}
& \score{54.90}
& \score{63.14}
& \score{89.13} \\

\rowcolor{scopeblue}
{\small\textbf{\textsc{SCOPD+}}}
& \score{2177.06}
& \score{59.60}
& \score{\underline{53.55}}
& \score{\textbf{43.65}}
& \score{\underline{60.30}}
& \score{\textbf{64.31}}
& \score{\textbf{95.25}} \\

\bottomrule
\end{tabular}
\end{adjustbox}

\end{minipage}%
\hfill%
\begin{minipage}[t]{0.485\textwidth}
\vspace{0pt}
\centering

\captionof{table}{
\textbf{Generalization across pruning operators.}
Trained only with VisionZip, \textsc{SCOPD+} generalizes to other pruners without adaptation.
}
\vspace{-0.5em}

\label{tab:ablate-pruning-methods}

\setlength{\tabcolsep}{1.5pt}
\renewcommand{\arraystretch}{1.15}

\begin{adjustbox}{max width=\linewidth}
\begin{tabular}{ll|rrrrrr|r}
\toprule
\benchhead{Pruner}
& \benchhead{Model}
& \benchhead{MME}
& \benchhead{MMStar}
& \benchhead{BLINK}
& \benchhead{VisOnly}
& \benchhead{MathVista}
& \benchhead{RWQA}
& \benchhead{Avg$_6$} \\
\midrule

\multirow{2}{*}{\small VisionZip}
& {\small Vanilla}
& \score{2040.13}
& \score{54.87}
& \score{48.97}
& \score{40.35}
& \score{55.30}
& \score{62.61}
& \score{88.74} \\

&
\cellcolor{scopeblue}{\small\textbf{\textsc{SCOPD+}}}
& \cellcolor{scopeblue}\score{\textbf{2177.06}}
& \cellcolor{scopeblue}\score{\textbf{59.60}}
& \cellcolor{scopeblue}\score{\textbf{53.55}}
& \cellcolor{scopeblue}\score{\textbf{43.65}}
& \cellcolor{scopeblue}\score{\textbf{60.30}}
& \cellcolor{scopeblue}\score{\textbf{64.31}}
& \cellcolor{scopeblue}\score{\textbf{95.25}} \\

\midrule

\multirow{2}{*}{\small DivPrune}
& {\small Vanilla}
& \score{1948.09}
& \score{52.27}
& \score{48.45}
& \score{38.52}
& \score{47.70}
& \score{58.04}
& \score{83.47} \\

&
\cellcolor{scopeblue}{\small\textbf{\textsc{SCOPD+}}}
& \cellcolor{scopeblue}\score{\textbf{2210.43}}
& \cellcolor{scopeblue}\score{\textbf{57.67}}
& \cellcolor{scopeblue}\score{\textbf{50.60}}
& \cellcolor{scopeblue}\score{\textbf{41.22}}
& \cellcolor{scopeblue}\score{\textbf{56.10}}
& \cellcolor{scopeblue}\score{\textbf{60.65}}
& \cellcolor{scopeblue}\score{\textbf{91.23}} \\

\midrule

\multirow{2}{*}{\small Random}
& {\small Vanilla}
& \score{1938.62}
& \score{50.07}
& \score{46.40}
& \score{41.57}
& \score{45.70}
& \score{54.90}
& \score{82.06} \\

&
\cellcolor{scopeblue}{\small\textbf{\textsc{SCOPD+}}}
& \cellcolor{scopeblue}\score{\textbf{2207.62}}
& \cellcolor{scopeblue}\score{\textbf{54.87}}
& \cellcolor{scopeblue}\score{\textbf{49.87}}
& \cellcolor{scopeblue}\score{\textbf{41.57}}
& \cellcolor{scopeblue}\score{\textbf{53.80}}
& \cellcolor{scopeblue}\score{\textbf{56.86}}
& \cellcolor{scopeblue}\score{\textbf{88.85}} \\

\midrule

\multirow{2}{*}{\small FastV}
& {\small Vanilla}
& \score{1954.39}
& \score{49.40}
& \score{47.45}
& \score{36.35}
& \score{49.60}
& \score{55.42}
& \score{81.47} \\

&
\cellcolor{scopeblue}{\small\textbf{\textsc{SCOPD+}}}
& \cellcolor{scopeblue}\score{\textbf{2096.13}}
& \cellcolor{scopeblue}\score{\textbf{51.53}}
& \cellcolor{scopeblue}\score{\textbf{49.55}}
& \cellcolor{scopeblue}\score{\textbf{38.78}}
& \cellcolor{scopeblue}\score{\textbf{51.80}}
& \cellcolor{scopeblue}\score{\textbf{58.82}}
& \cellcolor{scopeblue}\score{\textbf{86.03}} \\

\bottomrule
\end{tabular}
\end{adjustbox}

\end{minipage}
\vspace{-0.5em}
\end{table*}

\paragraph{Q1: Does visual sensitivity identify better distillation targets?}
We compare \textsc{SCOPD+} against dense \textsc{SCOPD} and alternative token-selection criteria, including random selection, teacher--student KL, TIP~\cite{xu2026tip}, and a negative control selecting the least visually sensitive positions. All selective variants retain $\rho=10\%$ of response tokens. Table~\ref{tab:ablate-sensitivity-metric} shows that \textsc{SCOPD+} achieves the best Avg$_6$ at $95.25\%$, outperforming dense \textsc{SCOPD} ($94.44\%$), Top-KL ($94.51\%$), TIP ($94.21\%$), and random selection ($93.56\%$), while using only $10\%$ of positions for distillation. Conversely, selecting the least sensitive positions drops performance to $89.13\%$, supporting visual sensitivity as an effective criterion for allocating supervision.

\vspace{-0.5em}

\paragraph{Q2: Does \textsc{SCOPD+} generalize across pruning operators?}
Although \textsc{SCOPD+} is trained only with VisionZip, we evaluate it without further adaptation under DivPrune~\cite{alvar2025divprune}, random pruning, and FastV~\cite{chen2024image}. As shown in Table~\ref{tab:ablate-pruning-methods}, \textsc{SCOPD+} improves Avg$_6$ across every tested pruner: from $88.74\%$ to $95.25\%$ with VisionZip, $83.47\%$ to $91.23\%$ with DivPrune, $82.06\%$ to $88.85\%$ with random pruning, and $81.47\%$ to $86.03\%$ with FastV. The transfer to FastV is especially informative because pruning occurs inside the language model, suggesting that the learned sparse-context adaptation is not specific to the VisionZip pruning mechanism.

\vspace{-0.5em}

\begin{table*}[t]
\centering

\begin{minipage}[t]{0.485\textwidth}
\centering
\captionof{table}{
\textbf{Generalization to Qwen3-VL-4B.}
Pruned models retain $10\%$ of visual tokens; Avg$_6$ is normalized to
Vanilla (100\%).
}

\vspace{-0.5em}

\label{tab:ablate-model-generalization}

\setlength{\tabcolsep}{1.8pt}
\renewcommand{\arraystretch}{1.20}

\begin{adjustbox}{max width=\linewidth}
\begin{tabular}{l|rrrrrr|r}
\toprule
\benchhead{Method}
& \benchhead{MME}
& \benchhead{MMStar}
& \benchhead{BLINK}
& \benchhead{VisOnly}
& \benchhead{MathVista}
& \benchhead{RWQA}
& \benchhead{Avg$_6$} \\
\midrule

{\small Vanilla (100\%)}
& \score{2403.47}
& \score{67.67}
& \score{65.75}
& \score{53.74}
& \score{69.90}
& \score{73.86}
& \score{100.00} \\

\midrule

{\small Vanilla (10\%)}
& \score{1999.68}
& \score{47.73}
& \score{51.76}
& \score{44.26}
& \score{42.50}
& \score{56.21}
& \score{75.29} \\

{\small \textsc{SCOPD}}
& \score{\underline{2261.94}}
& \score{\underline{52.93}}
& \score{\underline{54.29}}
& \score{\underline{44.87}}
& \score{\underline{47.30}}
& \score{\underline{61.70}}
& \score{\underline{81.60}} \\

\rowcolor{scopeblue}
{\small\textbf{\textsc{SCOPD+}}}
& \score{\textbf{2278.19}}
& \score{\textbf{53.87}}
& \score{\textbf{54.81}}
& \score{\textbf{44.96}}
& \score{\textbf{49.10}}
& \score{\textbf{63.40}}
& \score{\textbf{82.92}} \\

\bottomrule
\end{tabular}
\end{adjustbox}
\end{minipage}
\hfill
\begin{minipage}[t]{0.485\textwidth}
\centering
\captionof{table}{
\textbf{Image-to-video generalization} at $10\%$ retention.
No video-specific post-training is used.
}
\vspace{-0.5em}
\label{tab:ablate-video-generalization}

\setlength{\tabcolsep}{1.8pt}
\renewcommand{\arraystretch}{1.20}

\begin{adjustbox}{max width=\linewidth}
\begin{tabular}{l|rrrrr|r}
\toprule
\benchhead{Method}
& \benchhead{VideoMME}
& \benchhead{TempComp.}
& \benchhead{MVBench}
& \benchhead{MLVU}
& \benchhead{Video-TT}
& \benchhead{Avg$_5$} \\
\midrule

{\small Vanilla (100\%)}
& \score{54.00}
& \score{69.81}
& \score{62.98}
& \score{54.65}
& \score{36.20}
& \score{100.00} \\

\midrule

{\small Vanilla (10\%)}
& \score{49.37}
& \score{62.41}
& \score{56.60}
& \score{51.33}
& \score{32.50}
& \score{90.88} \\

{\small \textsc{SCOPD}}
& \score{\underline{52.78}}
& \score{\underline{65.44}}
& \score{\textbf{59.48}}
& \score{\underline{53.17}}
& \score{\underline{34.50}}
& \score{\underline{95.70}} \\

\rowcolor{scopeblue}
{\small\textbf{\textsc{SCOPD+}}}
& \score{\textbf{53.15}}
& \score{\textbf{66.01}}
& \score{\underline{59.45}}
& \score{\textbf{53.31}}
& \score{\textbf{35.50}}
& \score{\textbf{96.60}} \\

\bottomrule
\end{tabular}
\end{adjustbox}
\end{minipage}

\vspace{-1em}

\end{table*}

\paragraph{Q3: Do SCOPD and \textsc{SCOPD+} generalize to other reasoning VLMs?}
Our main experiments use Qwen2.5-VL-7B, so we additionally post-train Qwen3-VL-4B under the same $10\%$ sparse-context setup. Qwen3-VL uses stacked visual features from multiple vision layers; we therefore compute the VisionZip selection on the final-layer visual representation and apply the same selected token indices to the three intermediate visual streams. As shown in Table~\ref{tab:ablate-model-generalization}, pruning reduces Avg$_6$ to $75.29\%$, while \textsc{SCOPD} and \textsc{SCOPD+} recover it to $81.6\%$ and $82.92\%$, respectively. This suggests that sparse-context adaptation is not specific to the Qwen2.5-VL architecture.

\vspace{-0.5em}

\paragraph{Q4: Do SCOPD and \textsc{SCOPD+} generalize to videos?}
We next test whether adaptation learned entirely from image-based post-training transfers to video reasoning. Without any video-specific training, we evaluate the same LoRA-adapted models on five video benchmarks spanning general and temporal video understanding. Table~\ref{tab:ablate-video-generalization} shows that pruning reduces Avg$_5$ to $90.88\%$, while \textsc{SCOPD} recovers this to $95.70\%$ and \textsc{SCOPD+} further reaches $96.60\%$. Thus, the learned ability to use sparse visual context transfers across the image-to-video modality shift.

\vspace{-0.5em}

\begin{figure*}[t]
    \centering

    \begin{subfigure}[t]{0.485\textwidth}
        \centering
        \includegraphics[width=\linewidth]{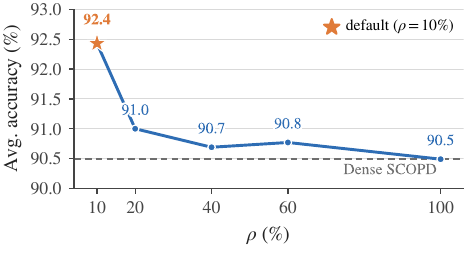}
        \caption{
        Effect of the supervision budget $\rho$
        }
        \label{fig:ablation-rho}
    \end{subfigure}
    \hfill
    \begin{subfigure}[t]{0.485\textwidth}
        \centering
        \includegraphics[width=\linewidth]{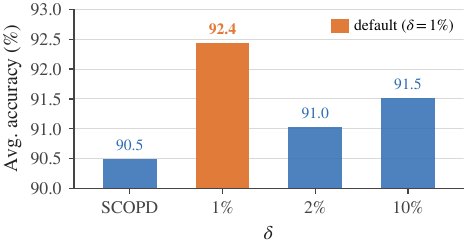}
        \caption{
        Effect of the visual intervention $\delta$
        }
        \label{fig:ablation-delta}
    \end{subfigure}

    \vspace{-0.4em}
    \caption{
\textbf{Sensitivity of \textsc{SCOPD+} to its hyperparameters.}
Results are normalized over 13 image benchmarks.
\textbf{(a)} Performance peaks at $\rho=10\%$, with no benefit from denser supervision.
\textbf{(b)} Performance is stable across visual-budget interventions $\delta$, with $\delta=1\%$ performing best overall.
}
    \label{fig:scopd-plus-design}
    \vspace{-0.5em}
\end{figure*}

\paragraph{Q5: How sensitive is \textsc{SCOPD+} to its hyperparameters ($\rho$ and $\delta$)?}
We study the two main hyperparameters of \textsc{SCOPD+}: the fraction of response positions selected for distillation, $\rho$, and the visual-budget intervention used to estimate sensitivity, $\delta$. As shown in Fig.~\ref{fig:scopd-plus-design}, performance is highest at $\rho=10\%$, while denser supervision provides no additional benefit. For the intervention, $\delta=1\%$ performs best. A small $\delta$ measures local sensitivity to marginal visual evidence near the deployment budget ($b=10\%$), whereas larger interventions compare against substantially richer visual representations and may emphasize positions that are less specific to the bottleneck at the operating point. We therefore use $\rho=10\%$ and $\delta=1\%$ by default.

\paragraph{Q6: How sensitive is \textsc{SCOPD+} to on-policy distillation design choices?}
We ablate the distillation divergence and teacher update strategy while keeping the remaining \textsc{SCOPD+} configuration fixed. Table~\ref{tab:opsd-design-ablation} shows that forward and reverse KL perform similarly ($92.43$ vs.\ $92.39$ Avg$_{13}$), while JSD is slightly weaker ($91.55$). Teacher regularization is more critical: freezing the teacher at the initial policy remains competitive ($91.37$), whereas using the current student weights directly as the teacher collapses to $12.35$. This mirrors Vision-OPD~\cite{yuan2026vision}, which likewise finds that an unregularized current-policy teacher collapses, motivating our EMA teacher. We therefore use forward KL with EMA by default. In Appendix~\ref{app:scopd-plus-gt}, we further show that providing the teacher with ground-truth supervision can substantially improve \textsc{SCOPD} when such labels are available.

\begin{table*}[t]
\centering
\caption{
\textbf{On-policy distillation design choices for \textsc{SCOPD+}.}
The default uses forward KL and an EMA teacher.
Avg$_{13}$ is the mean performance normalized to the unpruned Vanilla
model on each benchmark.
}
\label{tab:opsd-design-ablation}

\setlength{\tabcolsep}{2.3pt}
\renewcommand{\arraystretch}{1.14}

\begin{adjustbox}{max width=\textwidth}
\begin{tabular}{l|ccccccccccccc|c}
\toprule

\benchhead{Variant}
& \benchhead{MME}
& \benchhead{MMStar}
& \benchhead{MathVista}
& \benchhead{MathVerse}
& \benchhead{MMMU-Pro}
& \benchhead{Hallusion}
& \benchhead{CVBench}
& \benchhead{LogicVista}
& \benchhead{BLINK}
& \benchhead{VisOnly}
& \benchhead{HR4K}
& \benchhead{MME-RW-Lite}
& \benchhead{RWQA}
& \benchhead{Avg$_{13}$} \\
\midrule

\rowcolor{scopeblue}
{\small\textbf{\textsc{SCOPD+} (Default)}}
& \score{2177.06}
& \score{59.60}
& \score{60.30}
& \score{33.38}
& \score{42.95}
& \score{60.99}
& \score{71.96}
& \score{38.26}
& \score{53.55}
& \score{43.65}
& \score{66.00}
& \score{43.15}
& \score{64.31}
& \score{\textbf{92.43}} \\

\midrule
\rowcolor{blockgray}
\multicolumn{15}{c}{
    \small\textit{Distillation Divergence}
} \\

{\small Reverse KL}
& \score{2215.71}
& \score{60.47}
& \score{60.20}
& \score{32.61}
& \score{43.18}
& \score{61.83}
& \score{72.17}
& \score{34.90}
& \score{54.60}
& \score{43.39}
& \score{66.50}
& \score{43.20}
& \score{65.10}
& \score{\underline{92.39}} \\

{\small JSD ($\beta=0.5$)}
& \score{2198.50}
& \score{59.87}
& \score{60.90}
& \score{34.01}
& \score{41.16}
& \score{61.72}
& \score{71.79}
& \score{34.45}
& \score{53.76}
& \score{42.61}
& \score{63.62}
& \score{43.41}
& \score{64.84}
& \score{91.55} \\

\midrule
\rowcolor{blockgray}
\multicolumn{15}{c}{
    \small\textit{Teacher Strategy}
} \\

{\small Fixed Teacher}
& \score{2213.41}
& \score{59.27}
& \score{58.90}
& \score{32.11}
& \score{41.45}
& \score{60.25}
& \score{71.52}
& \score{36.02}
& \score{53.76}
& \score{42.70}
& \score{65.25}
& \score{43.04}
& \score{66.27}
& \score{91.37} \\

{\small Shared}
& \score{797.76}
& \score{9.07}
& \score{2.80}
& \score{0.00}
& \score{14.05}
& \score{3.36}
& \score{1.41}
& \score{12.98}
& \score{0.53}
& \score{0.17}
& \score{11.38}
& \score{7.24}
& \score{5.49}
& \score{12.35} \\

\bottomrule
\end{tabular}
\end{adjustbox}

\end{table*}

\paragraph{Q7: What is the compute overhead of \textsc{SCOPD+}?}
The visual-budget intervention is used only during post-training and introduces no additional inference-time model computation. On the 500-example diagnostic set, SCOPD and SCOPD+ generate 153.0 and 155.7 tokens on average, respectively, comparable to 155.3 for the unpruned Vanilla model. During training, SCOPD+ adds 21.4\% theoretical compute over SCOPD but only 1.9\% measured time, with essentially unchanged peak memory. Full measurements are provided in Appendix~\ref{app:compute}.
\section{Conclusion}
\label{sec:conclusion}

We identified a \emph{representation--utilization gap} in reasoning VLMs under visual-token pruning: sparse representations can retain sufficient evidence for correct solutions, yet the language model may fail to use it reliably. Motivated by this observation, we introduced \textsc{SCOPD}, which adapts models to sparse visual contexts through on-policy self-distillation from a privileged full-context teacher, and \textsc{SCOPD+}, which focuses supervision on visually sensitive reasoning positions identified through a small visual-budget intervention. Across token budgets, pruning operators, model families, and image and video benchmarks, both methods improved sparse-context reasoning without architectural changes or additional inference-time computation.

\paragraph{Limitations and future work.}
Our experiments deliberately focused on controlled sparse-context adaptation, using VisionZip as the primary training-time pruning operator and fixed retention budgets. Although the learned adaptation transferred to other pruning methods, future work could study a broader range of in-LLM and dynamic pruning strategies, alternative pruning schedules, and jointly learned compression policies. We also primarily evaluated short-form image and video reasoning; extending sparse-context adaptation to longer-horizon video, embodied, and agentic tasks may reveal additional challenges in retaining and utilizing visual evidence over time.

\subsection*{Reproducibility Statement}
We provide detailed training and evaluation configurations, data processing,
prompts, pruning settings, baseline implementations, and compute measurements
in the appendix. The fixed-context Pass@$K$ protocol, validity-judge prompt,
GRPO setup, and additional ablations are also documented to facilitate
reproduction of our results. We additionally include our training code in the
supplementary material accompanying this submission.

\subsection*{AI Use Statement}
In this work, generative AI tools were used to provide feedback on research
methodology and experimental design, assist with interpretation of experimental
results, and support drafting and editing of the manuscript. They were also
used for code and figure/table assistance where applicable. All AI-assisted
outputs, including methodological suggestions, code, analyses, citations, and
written text, were reviewed and verified by the authors. The authors take
responsibility for the final content of the paper and all reported results.

\newpage

\newpage
\appendix
\section{Expanded Fixed-Context Pass@\texorpdfstring{$K$}{K} Analysis}
\label{app:pass-at-k}

\begin{figure*}[t]
    \centering
    \includegraphics[width=\linewidth]
    {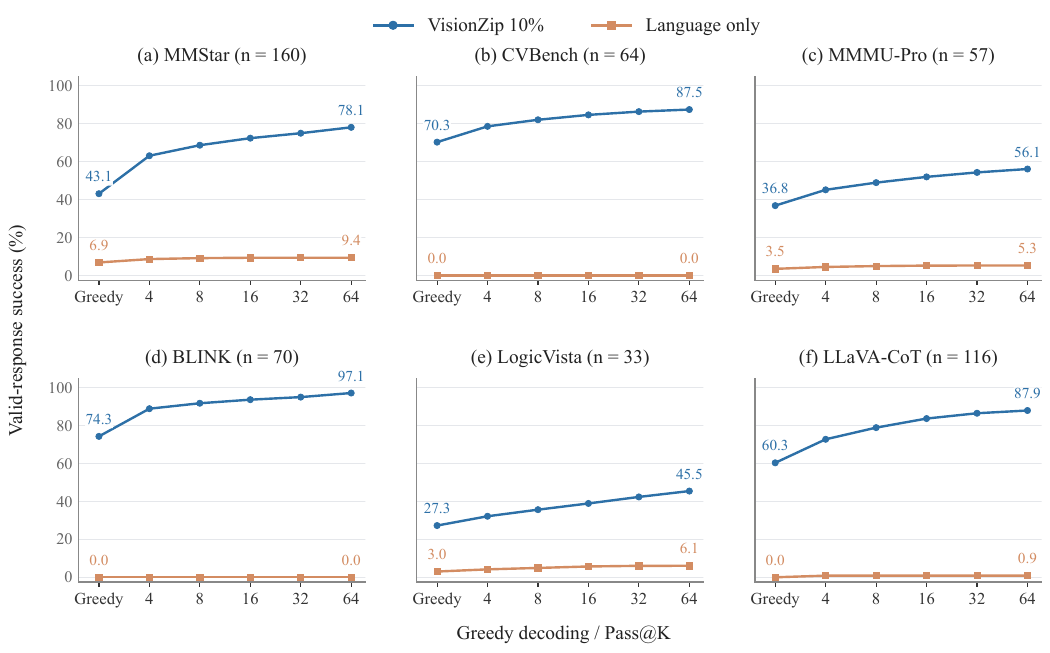}
    \caption{
    \textbf{Recoverability under aggressive visual-token pruning.}
    We report greedy and Pass@$K$ performance
    ($K\in\{4,8,16,32,64\}$) across the six sources in our
    500-example diagnostic set. VisionZip retains $10\%$ of visual tokens,
    while the language-only control receives no visual input. Performance
    consistently improves with sampling under the fixed sparse context,
    whereas the language-only control remains near zero.
    }
    \label{fig:passk-benchmark-recovery}
\end{figure*}

\paragraph{Diagnostic set.}
We construct a frozen diagnostic set of 500 free-form numerical
image--question pairs from MMStar, CVBench, MMMU-Pro, BLINK, LogicVista,
and LLaVA-CoT, containing 493 distinct images. For multiple-choice datasets,
we remove the answer options only when the remaining question has an
independently meaningful numerical answer. The reference answer is retained
for evaluation but is never provided during generation. The set contains
160, 64, 57, 70, 33, and 116 examples from the six sources, respectively,
and should be viewed as a controlled diagnostic rather than official
benchmark evaluation. The cohort is selected from examples associated with successful
unpruned greedy generations.

\paragraph{Generation setup.}
We use the unadapted Qwen2.5-VL-7B-Instruct model throughout this analysis.
For the sparse condition, VisionZip retains $10\%$ of visual tokens
($5\%$ dominant and $5\%$ contextual tokens with contextual merging).
For each example, the pruned visual representation is computed once and
held fixed across all generations. Thus, repeated samples differ only in
the language-generation trajectory, not in which visual tokens are retained.

We generate 64 stochastic responses per example using temperature $0.7$,
top-$p=0.95$, top-$k=50$, a maximum of 1024 new tokens, and no repetition
penalty. Greedy responses are generated separately with sampling disabled.
We additionally evaluate a language-only control using the identical textual
input and decoding setup but with the image and visual placeholder removed.

\paragraph{Prompt.}
All conditions use the default system message
\texttt{You are a helpful assistant.}
For each question, we append the same reasoning instruction used in our
main evaluation:
\begin{quote}
\small
\texttt{\{question\}}\\
\texttt{First output the thinking process in <think> </think> tags and then output}\\
\texttt{the final answer in <answer> </answer> tags.}
\end{quote}
The language-only control uses exactly the same textual prompt.

\begin{figure*}[t]
    \centering

    \includegraphics[width=\linewidth]
    {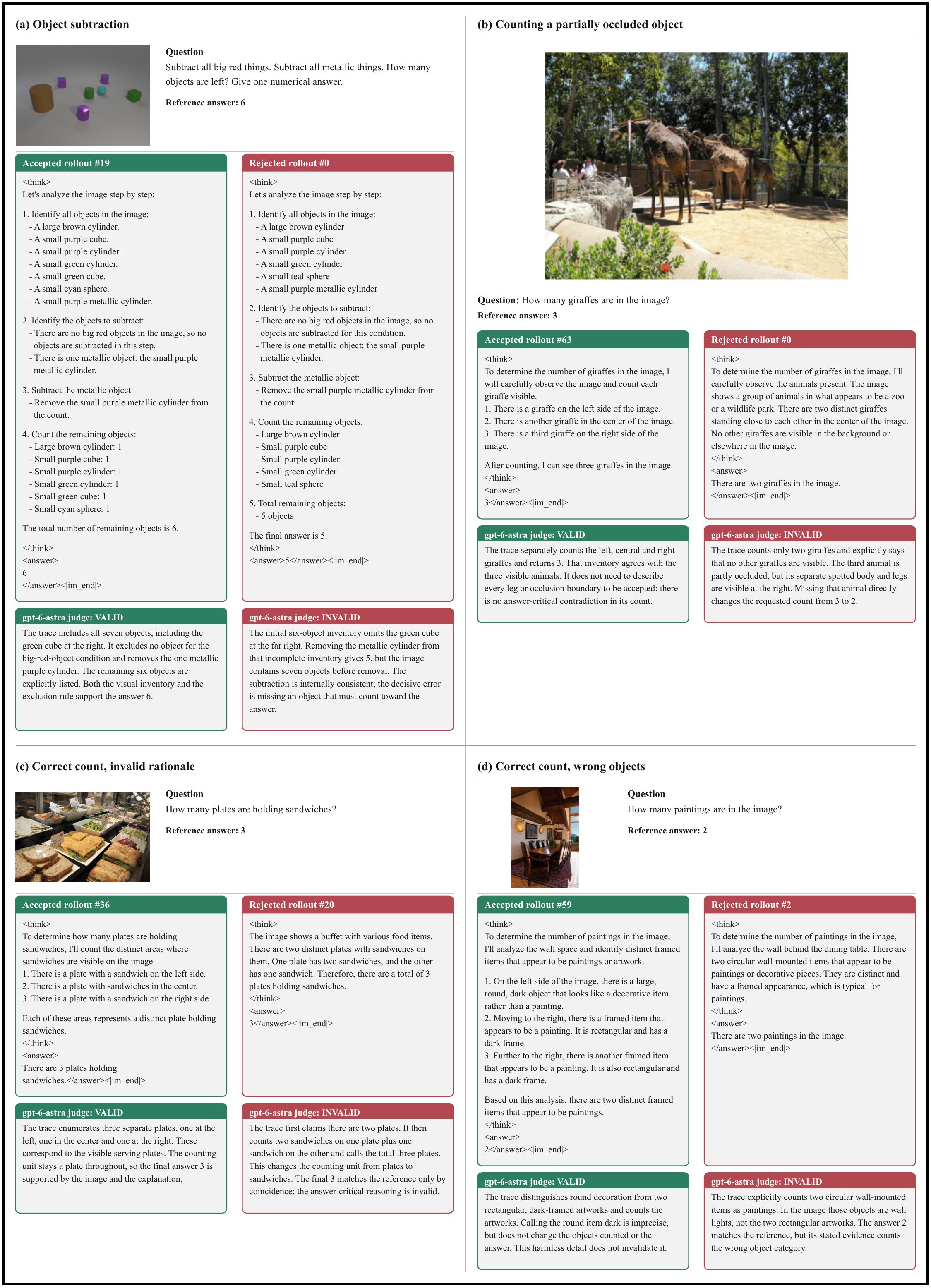}
    
    \caption{
    \textbf{Examples of reasoning-aware validity judgments.}
    Four sampled trajectories illustrate cases where answer correctness alone can miss unsupported or visually inconsistent reasoning, motivating our GPT-based validity judge.
    }
        
    \label{fig:passk-judge-examples}
\end{figure*}

\paragraph{Reasoning-aware validity evaluation.}
Answer matching alone can overestimate recovery when the final answer is correct but the reasoning relies on an incorrect visual observation, invalid operation, or unsupported guess. We therefore use \texttt{gpt-6-astra} with reasoning effort set to \texttt{max} as a semantic validity judge. For each generation, the judge receives the original image, question, reference answer, and complete model response, and verifies both answer correctness and whether the decisive observations and operations are supported by the available evidence. Unsupported guesses and answer-critical hallucinations are rejected, while uncertain cases are conservatively counted as failures. Fig.~\ref{fig:passk-judge-examples} illustrates representative successful and failed trajectories, including cases where the final answer alone would not reliably indicate whether the reasoning is valid.

The judge prompt is:
\begin{quote}
\small
Inspect the original image, question, reference answer, and complete response.

Accept a correct answer whose decisive facts and operations are supported.

Allow errors or omissions that do not affect the answer or its essential basis.

Reject answer-critical hallucinations, wrong operations, and unsupported guesses.

Do not reject text-only solutions if the question itself supplies enough evidence.

If validity cannot be established, retain an \texttt{uncertain} label.
\end{quote}

\paragraph{Pass@$K$ metric.}
Let $c_i$ denote the number of valid responses among the 64 stochastic
generations for example $i$. We use the standard Pass@$K$ estimator~\cite{chen2021evaluating}:
\begin{equation}
\widehat{\mathrm{Pass@}K}
=
\frac{1}{N}
\sum_{i=1}^{N}
\left[
1 -
\frac{\binom{64-c_i}{K}}
     {\binom{64}{K}}
\right].
\label{eq:passk}
\end{equation}
Greedy accuracy is computed from a separately generated deterministic
response and is therefore distinct from sampled Pass@1. We additionally
report $\mathrm{Hit}_{\geq4}@64$, the fraction of examples for which at
least four of the 64 generations are valid, as a stricter measure of
repeated recoverability.

\paragraph{Results and robustness.}
As shown in Figs.~\ref{fig:recoverability-b} and ~\ref{fig:passk-benchmark-recovery}, with the visual representation fixed at VisionZip $10\%$, performance rises from $53.20\%$ under greedy decoding to $79.60\%$ at Pass@64, while $\mathrm{Hit}_{\geq4}@64$ reaches $74.20\%$. In contrast, the language-only control increases only from $2.80\%$ to $4.20\%$. These results support our central diagnostic claim: many failures under aggressive pruning remain recoverable from the \emph{same} sparse visual representation rather than requiring a different pruning outcome.

\section{Additional Analysis of Visually Sensitive Tokens}
\label{app:visual-sensitivity-samples}

As discussed in Sections~\ref{sec:SCOPD} and~\ref{sec:SCOPD-plus}, large
teacher--student KL divergence does not necessarily indicate that a reasoning
position depends on visual evidence. In practice, high-KL positions can arise
from purely language-based variation, such as capitalization, paraphrases,
alternative but equivalent wording, or other response-format differences. In
such cases, the student and teacher may disagree strongly even though the
underlying visual evidence is not the source of the discrepancy. Similar studies have been done on language modeling~\cite{armandpour2026unmasking}.

\begin{figure*}[t]
    \centering
    \includegraphics[width=\linewidth]
    {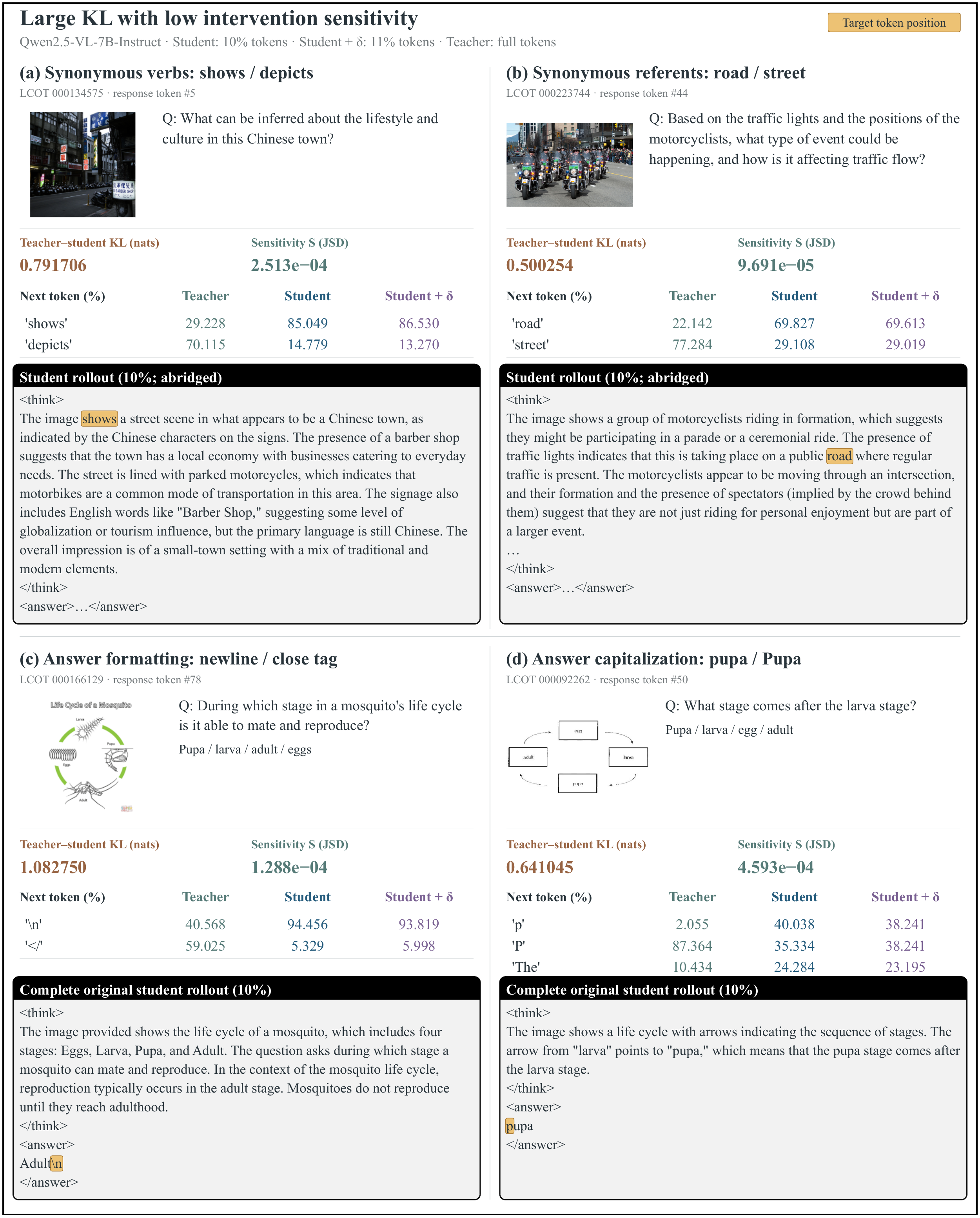}
    \caption{
\textbf{High teacher--student KL does not always imply visual relevance.}
Language-level differences can produce large KL despite low visual sensitivity.
}
    \label{fig:high-kl-low-sensitivity}
\end{figure*}

Figure~\ref{fig:high-kl-low-sensitivity} provides additional examples of this
phenomenon. Across these samples, some positions have high teacher--student KL
but remain largely unchanged under our visual-budget intervention, indicating
low visual sensitivity. These cases illustrate why disagreement alone can be a
noisy proxy for useful visual supervision. By contrast, our intervention-based
sensitivity signal is more selective: it highlights positions whose predictions
actually respond to additional visual evidence, and therefore better matches
the goal of identifying where privileged visual supervision is most useful.

\section{Training and Evaluation Details}
\label{app:eval-train-setup}

This section provides the full training, data, and evaluation configuration used throughout our experiments.

\paragraph{Base models and visual-token pruning.}
Unless otherwise stated, all experiments use Qwen2.5-VL-7B-Instruct~\cite{bai2025qwen25vltechnicalreport}. Our default pruning operator is VisionZip~\cite{yang2025visionzip}, with a training-time retention budget of $b=10\%$. VisionZip combines attention-derived token importance with contextualized visual features. In our ablations, we additionally evaluate DivPrune~\cite{alvar2025divprune} as a diversity-based pruning strategy, Random pruning as a control, and FastV~\cite{chen2024image} as a representative in-LLM, query-aware pruning method. We also study transfer to Qwen3-VL-4B~\cite{bai2025qwen3}. Unless otherwise specified, the vision encoder and multimodal projector remain frozen, and only the language model is adapted.

\paragraph{VisionZip.}
VisionZip is a training-free and text-agnostic compression method that reduces redundancy in the visual sequence before it is consumed by the language model.
Given the visual tokens produced by the vision encoder, VisionZip first identifies a set of \emph{dominant tokens} according to their self-attention importance.
For vision encoders without a dedicated \texttt{[CLS]} token, token importance is estimated by the average attention received from the other visual tokens.
The highest-scoring tokens are retained directly, as they tend to aggregate a large fraction of the visual information.
The remaining non-dominant tokens are then grouped according to similarity in the vision encoder's key-feature space and merged into a smaller set of \emph{contextual tokens}, preserving complementary information that may not be captured by the dominant tokens.
The resulting dominant and contextual tokens together form the compressed visual sequence supplied to the multimodal projector and language model.

We denote the visual-token retention ratio by $b$, defined relative to the visual sequence after image preprocessing and the 1280-token cap.
For example, $b=10\%$ retains approximately one tenth of the available visual tokens, whereas $b=100\%$ corresponds to the unpruned visual context.
Unless otherwise specified, training is performed at $b=10\%$, while evaluation additionally considers larger token budgets to measure robustness across pruning levels.
Because VisionZip performs token selection using only the visual representation and does not depend on the question or generated text, the retained visual context remains fixed throughout an autoregressive reasoning trajectory.

\paragraph{Training data.}
We post-train on 10K examples derived from LLaVA-CoT~\cite{xu2025llava}. We use the processed version introduced by OpenMMReasoner~\cite{zhang2026openmmreasoner}, which converts the original data into a consistent reasoning format with explicit
\texttt{<think>...</think>} and \texttt{<answer>...</answer>} fields. During training, images are capped at 1280 visual tokens before pruning.
The dataset provides image--question pairs together with reference answers and reasoning traces, which are used by supervised baselines such as SFT and EPIC~\cite{wen2026efficient} when required. In contrast, \textsc{SCOPD} and \textsc{SCOPD+} use only the input image and question; supervision is obtained online from the full-context teacher and therefore requires no ground-truth answers or reasoning traces.

\paragraph{Reasoning format.}
All post-training and evaluation runs use the same reasoning format. The model is prompted to produce a reasoning trace enclosed by
\texttt{<think>...</think>}, followed by its final prediction inside
\texttt{<answer>...</answer>}.
Accordingly, a ``trajectory'' in our formulation refers to the autoregressively generated reasoning sequence together with the final answer. For \textsc{SCOPD} and \textsc{SCOPD+}, these trajectories are generated on-policy by the sparse-context student, while the full-context teacher scores the same student-generated prefixes.

\paragraph{Post-training configuration.}
All methods use LLM-only LoRA with rank $r=16$, scaling factor $\alpha=32$, and zero dropout. LoRA adapters are applied to the attention and feed-forward projections of all language-decoder layers, while the vision encoder and multimodal projector remain frozen. Unless otherwise specified, we use a constant learning rate of $2\times10^{-5}$ and an effective batch size of 32 across four GPUs. All methods are trained from the same Qwen2.5-VL-7B-Instruct initialization on the same training examples.

SFT, \textsc{SCOPD}, and \textsc{SCOPD+} use VisionZip~\cite{yang2025visionzip} at $10\%$ visual-token retention during training. EPIC~\cite{wen2026efficient} follows its original progressive retention schedule and combined supervised/distillation objective under the same base model and LoRA configuration. For \textsc{SCOPD} and \textsc{SCOPD+}, the teacher is an exponential moving average of the student parameters with decay $0.9999$. For \textsc{SCOPD+}, we retain the top $\rho=0.1$ fraction of response positions according to visual sensitivity. The intervention context is constructed by slightly increasing the available visual-token budget, as described in Section~\ref{sec:SCOPD-plus}. LoRA adapters are merged into the base model before evaluation.

\paragraph{Baselines.}
We compare against baselines that isolate different ways of adapting a VLM to sparse visual context.
\textbf{Vanilla} is the pretrained model evaluated under pruning without post-training, measuring degradation from visual-token compression alone.
\textbf{SFT} is fine-tuned on the same $10\%$ sparse visual context using reference reasoning traces and answers, testing whether conventional supervised adaptation is sufficient.
\textbf{EPIC}~\cite{wen2026efficient} is a compression-aware baseline based on progressive consistency distillation.
\textbf{GRPO} is an RLVR-style on-policy baseline that generates groups of reasoning trajectories from the same sparse visual context and optimizes them using outcome-based rewards; full training details are provided in Appendix~\ref{app:grpo-details}.

For the token-selection ablations, we additionally compare against random selection, positions with the largest teacher--student KL divergence, TIP~\cite{xu2026tip}, and the lowest-sensitivity positions as a negative control. These baselines test whether the gains of \textsc{SCOPD+} arise from sparse backpropagation, generic teacher--student disagreement, or specifically from sensitivity to visual context.

\paragraph{Image benchmarks.}
Our main image evaluation comprises 13 benchmarks spanning complementary vision--language capabilities:
MME~\cite{fu2025mme},
MMStar~\cite{chen2024we},
MathVista~\cite{lu2023mathvista},
MathVerse-VO~\cite{zhang2024mathverse},
MMMU-Pro~\cite{yue2025mmmu},
HallusionBench~\cite{guan2024hallusionbench},
CVBench~\cite{tong2024cambrian},
LogicVista~\cite{xiao2024logicvista},
BLINK~\cite{fu2024blink},
VisOnlyQA~\cite{kamoi2024visonlyqa},
HR4K~\cite{wang2025divide},
RealWorldQA~\cite{xai2024grok15v},
and MME-Realworld-Lite~\cite{zhang2025mme}.
Together, these benchmarks cover general perception, hallucination, spatial and fine-grained visual understanding, mathematical and logical reasoning, high-resolution perception, and real-world visual reasoning.

\paragraph{Video benchmarks.}
To test whether sparse-context adaptation extends beyond static images, we additionally evaluate on five video benchmarks:
VideoMME~\cite{fu2025video},
MLVU~\cite{zhou2025mlvu},
TempCompass-MCQ~\cite{liu2024tempcompass},
MVBench~\cite{li2024mvbench},
and Video-TT~\cite{zhang2025towards}.
These benchmarks cover general video understanding, temporal reasoning, long-form video understanding, and video perception.

\paragraph{Evaluation protocol.}
All methods are evaluated with the same reasoning prompt and decoding configuration using VLMEvalKit~\cite{duan2024vlmevalkit}. For image benchmarks, inputs are capped at 4096 visual tokens. For video benchmarks, we uniformly sample 32 frames and allow up to 12,288 visual tokens per video. We use greedy decoding and append the following reasoning-inducing instruction to each question:
\begin{quote}
\small
\texttt{\{question\}}\\
\texttt{First output the thinking process in <think> </think> tags and then output}\\
\texttt{the final answer in <answer> </answer> tags.}
\end{quote}

After generation, we extract the content of the
\texttt{<answer>...</answer>} field and compare it with the benchmark reference. Because reasoning models can produce semantically equivalent answers with different surface forms, we use Qwen3.6-27B~\cite{qwen3.6-27b} as an answer-matching judge when direct matching is ambiguous. The judge receives the question, reference answer, and model prediction and determines whether the final answer is semantically correct. It is used only for answer normalization and correctness matching, not to evaluate the generated reasoning trace.

\section{Extreme Compression at 5\% Visual-Token Retention}
\label{app:aggressive-pruning}

To test whether sparse-context adaptation remains effective beyond the pruning budgets considered in the main experiments, we evaluate an extreme setting that retains only $5\%$ of the original visual tokens. Table~\ref{tab:five-percent-retention} reports results on eight representative image benchmarks, with Avg$_8$ normalized to the corresponding unpruned Vanilla performance.

At this budget, VisionZip substantially degrades the Vanilla model, reducing Avg$_8$ to $75.99\%$. \textsc{SCOPD} recovers the aggregate to $83.63\%$, while \textsc{SCOPD+} reaches $83.66\%$. The improvements are broad across benchmarks: \textsc{SCOPD} is particularly effective on MME, MMStar, MathVista, and VisOnlyQA, while \textsc{SCOPD+} performs best on HallusionBench, CVBench, LogicVista, and RealWorldQA. These results show that adapting the language model remains beneficial even when $95\%$ of the visual tokens are removed, although the remaining gap to the unpruned model also highlights the limits of recovering from increasingly severe information loss.

\begin{table*}[t]
\centering
\caption{
\textbf{Performance under extreme visual-token pruning.}
Results at $5\%$ token retention; Avg$_8$ is normalized to the unpruned Vanilla model.
}
\label{tab:five-percent-retention}

\setlength{\tabcolsep}{3.0pt}
\renewcommand{\arraystretch}{1.20}

\begin{adjustbox}{max width=\textwidth}
\begin{tabular}{l|c|cccccccc|c}
\toprule
\benchhead{Method}
& \benchhead{Retention}
& \benchhead{MME}
& \benchhead{MMStar}
& \benchhead{MathVista}
& \benchhead{Hallusion}
& \benchhead{CVBench}
& \benchhead{LogicVista}
& \benchhead{VisOnlyQA}
& \benchhead{RealWorldQA}
& \benchhead{Avg$_8$} \\
\midrule

{\small Vanilla}
& \score{100\%}
& \score{2250.63}
& \score{64.27}
& \score{66.70}
& \score{66.25}
& \score{75.19}
& \score{47.87}
& \score{44.70}
& \score{64.97}
& \score{100.00\%} \\

\midrule

{\small Vanilla}
& \score{5\%}
& \score{1789.51}
& \score{46.53}
& \score{39.40}
& \score{51.10}
& \score{63.53}
& \score{31.54}
& \score{39.30}
& \score{52.94}
& \score{75.99\%} \\

{\small SCOPD \textbf{(Ours)}}
& \score{5\%}
& \score{\textbf{2086.28}}
& \score{\textbf{52.80}}
& \score{\textbf{50.20}}
& \score{\underline{55.63}}
& \score{\underline{67.00}}
& \score{\underline{33.56}}
& \score{\textbf{40.96}}
& \score{\underline{54.64}}
& \score{\underline{83.63\%}} \\

\rowcolor{scopeblue}
{\small\textbf{SCOPD+ (Ours)}}
& \score{5\%}
& \score{\underline{2076.02}}
& \score{\underline{52.00}}
& \score{\underline{47.70}}
& \score{\textbf{57.20}}
& \score{\textbf{68.28}}
& \score{\textbf{34.23}}
& \score{\underline{40.35}}
& \score{\textbf{55.69}}
& \score{\textbf{83.66\%}} \\

\bottomrule
\end{tabular}
\end{adjustbox}

\end{table*}

\begin{figure*}[t]
    \centering

    \begin{subfigure}[t]{0.485\textwidth}
        \centering
        \includegraphics[width=\linewidth]{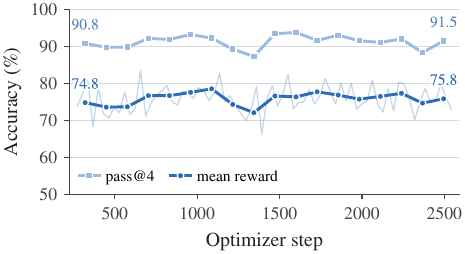}
        \caption{Accuracy of sampled responses.}
        \label{fig:grpo-reward}
    \end{subfigure}
    \hfill
    \begin{subfigure}[t]{0.485\textwidth}
        \centering
        \includegraphics[width=\linewidth]{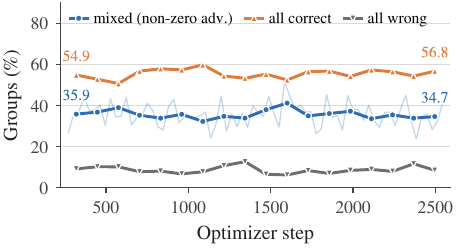}
        \caption{Composition of the $G=4$ groups.}
        \label{fig:grpo-groups}
    \end{subfigure}

    \vspace{0.3em}

    \begin{subfigure}[t]{0.485\textwidth}
        \centering
        \includegraphics[width=\linewidth]{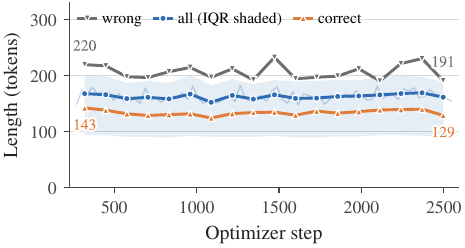}
        \caption{Response length.}
        \label{fig:grpo-length}
    \end{subfigure}
    \hfill
    \begin{subfigure}[t]{0.485\textwidth}
        \centering
        \includegraphics[width=\linewidth]{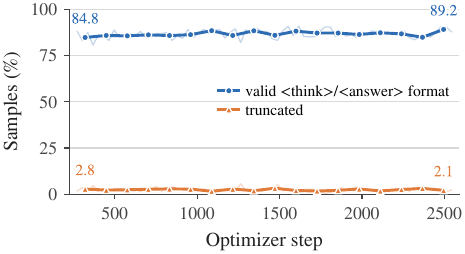}
        \caption{Format compliance and truncation.}
        \label{fig:grpo-format}
    \end{subfigure}

    \vspace{-0.4em}
    \caption{
\textbf{Training dynamics of the GRPO baseline}.
Each point is a 128-step window; faint lines are 32-step windows.
Accuracy is exact match on the prompts with a checkable answer (multiple choice, numeric, yes/no; 65\% of prompts).
\textbf{(a)} Mean accuracy and Pass@4 are flat at $\approx$76\% and $\approx$91\%.
\textbf{(b)} Only $\approx$35\% of groups contain both correct and incorrect responses and thus receive a non-zero advantage; $\approx$55\% are already all correct.
\textbf{(c)} Mean length stays at $\approx$160 tokens (shaded: interquartile range); incorrect responses are consistently longer than correct ones.
\textbf{(d)} Format compliance improves slightly (84.8\% $\rightarrow$ 89.2\%); truncation stays below 3\%.
}
    \label{fig:grpo-baseline}
    \vspace{-0.5em}
\end{figure*}

\section{GRPO discussions}
\label{app:grpo-details}
We train the same base model with GRPO~\cite{shao2024deepseekmath} on the training prompts, using the R1-style
\texttt{<think>}/\texttt{<answer>} format.
Each optimizer step samples $G=4$ responses for one prompt per GPU on 4 GPUs (16 responses per step);
rewards are assigned by a Qwen2.5-7B judge, and advantages are normalized within each group.
A 2{,}500-step run takes about 30 hours.
Figure~\ref{fig:grpo-baseline} shows the training dynamics.
Accuracy, Pass@4, and response length do not change over training, and only format compliance improves.
The reason is visible in Figure~\ref{fig:grpo-groups}: with $G=4$, the base model already answers all four samples correctly for over half of the prompts, so these groups have zero advantage and contribute no gradient.
The learning signal is confined to the roughly one third of groups with mixed outcomes, which is too little to move the policy within our compute budget.
Larger groups or harder prompts could raise the fraction of informative groups, but at a proportionally higher cost per step.

\section{Ground-Truth-Augmented Distillation}
\label{app:scopd-plus-gt}

\paragraph{Ground-truth supervision for the teacher.}
Our default \textsc{SCOPD} and \textsc{SCOPD+} require no ground-truth answers or reasoning traces: the sparse-context student generates trajectories on-policy, and the full-context teacher provides token-level supervision on the same prefixes. We additionally test whether privileged ground-truth information can further strengthen this supervision when labels are available. In this variant, the student rollout remains unchanged, but the full-context teacher is additionally provided with the ground-truth answer when scoring the student-generated trajectory. We evaluate this augmentation with \textsc{SCOPD} only and do not combine it with the selective \textsc{SCOPD+} objective.

Table~\ref{tab:scopd-gt} shows that ground-truth-augmented \textsc{SCOPD} improves Avg$_{13}$ from $90.49$ to $92.86$ at $10\%$ visual-token retention. The improvement also persists with the full visual context, where performance increases from $100.67$ to $102.43$. Thus, privileged ground-truth answers can provide complementary supervision beyond full visual context alone. Importantly, this variant changes the supervision assumptions and is therefore not used as our default: the main \textsc{SCOPD} and \textsc{SCOPD+} results remain fully independent of ground-truth answers or reasoning traces.

\begin{table*}[t]
\centering
\caption{
\textbf{Ground-truth-augmented \textsc{SCOPD}.}
The full-context teacher is additionally provided with the ground-truth answer, while student trajectories remain on-policy. Avg$_{13}$ is normalized to the unpruned Vanilla model.
}
\label{tab:scopd-gt}

\setlength{\tabcolsep}{2.3pt}
\renewcommand{\arraystretch}{1.14}

\begin{adjustbox}{max width=\textwidth}
\begin{tabular}{l|ccccccccccccc|c}
\toprule
\benchhead{Method}
& \benchhead{MME}
& \benchhead{MMStar}
& \benchhead{MathVista}
& \benchhead{MathVerse}
& \benchhead{MMMU-Pro}
& \benchhead{Hallusion}
& \benchhead{CVBench}
& \benchhead{LogicVista}
& \benchhead{BLINK}
& \benchhead{VisOnly}
& \benchhead{HR4K}
& \benchhead{MME-RW}
& \benchhead{RWQA}
& \benchhead{Avg$_{13}$} \\
\midrule

\rowcolor{blockgray}
\multicolumn{15}{c}{
    \small\textit{Full Visual Context (100\% Tokens)}
} \\

{\small \textsc{SCOPD}}
& \score{2322.58}
& \score{64.20}
& \score{68.00}
& \score{43.02}
& \score{45.49}
& \score{\textbf{68.14}}
& \score{73.20}
& \score{\textbf{47.20}}
& \score{56.71}
& \score{\textbf{45.65}}
& \score{64.62}
& \score{47.68}
& \score{\textbf{67.58}}
& \score{100.67} \\

{\small \textsc{SCOPD} + GT}
& \score{\textbf{2434.38}}
& \score{\textbf{66.27}}
& \score{\textbf{69.00}}
& \score{\textbf{43.40}}
& \score{\textbf{47.57}}
& \score{67.09}
& \score{\textbf{77.46}}
& \score{44.30}
& \score{\textbf{56.97}}
& \score{44.09}
& \score{\textbf{69.25}}
& \score{\textbf{50.65}}
& \score{67.32}
& \score{\textbf{102.43}} \\

\midrule
\rowcolor{blockgray}
\multicolumn{15}{c}{
    \small\textit{Retain 10\% Visual Tokens}
} \\

{\small \textsc{SCOPD}}
& \score{2201.28}
& \score{58.60}
& \score{\textbf{60.60}}
& \score{\textbf{30.46}}
& \score{42.95}
& \score{\textbf{60.46}}
& \score{72.65}
& \score{34.68}
& \score{52.81}
& \score{42.09}
& \score{63.75}
& \score{41.84}
& \score{64.31}
& \score{90.49} \\

{\small \textsc{SCOPD} + GT}
& \score{\textbf{2318.36}}
& \score{\textbf{60.67}}
& \score{\textbf{60.60}}
& \score{29.44}
& \score{\textbf{43.76}}
& \score{59.41}
& \score{\textbf{74.17}}
& \score{\textbf{38.26}}
& \score{\textbf{52.92}}
& \score{\textbf{42.35}}
& \score{\textbf{68.38}}
& \score{\textbf{44.76}}
& \score{\textbf{65.10}}
& \score{\textbf{92.86}} \\

\bottomrule
\end{tabular}
\end{adjustbox}
\end{table*}

\section{Compute}
\label{app:compute}

\paragraph{Compute and memory measurement.}
We measure the additional cost introduced by the visual-budget intervention in
\textsc{SCOPD+} relative to dense \textsc{SCOPD}. Measurements are collected
on a single NVIDIA L40 GPU using a batch size of 32. FLOPs and wall-clock time
are normalized per sample, while peak GPU memory denotes the maximum memory
allocated during the batched run.

\textsc{SCOPD} requires 54.25 TFLOPs and 17.21\,s per sample, with a peak GPU
memory of 25.02\,GiB. \textsc{SCOPD+} increases compute to 65.84 TFLOPs per
sample ($+21.4\%$), reflecting the additional forward pass used to estimate
visual sensitivity. Despite this increase in theoretical compute, measured
wall-clock time rises only to 17.54\,s per sample ($+1.9\%$), while peak memory
changes negligibly to 25.13\,GiB ($+0.4\%$). Thus, the intervention adds
moderate training-time compute but little additional latency or peak memory
in our implementation. With our LoRA setup, full post-training on four NVIDIA
L40 GPUs takes approximately 14 hours.

\paragraph{Inference cost and response length.}
Both \textsc{SCOPD} and \textsc{SCOPD+} are post-training methods: the
visual-budget intervention and privileged teacher are removed at inference,
and the deployed model uses the same architecture and sparse visual context
as the underlying pruned VLM. To check whether adaptation indirectly increases
decoding cost by producing substantially longer reasoning trajectories, we
measure the mean number of generated tokens over the 500 examples used in our
fixed-context Pass@$K$ analysis. The unpruned Vanilla model generates 155.27
tokens on average, compared with 153.00 for \textsc{SCOPD} and 155.70 for
\textsc{SCOPD+}. Thus, our methods do not induce longer-than-base reasoning
trajectories. For reference, VisionZip at $10\%$ retention generates 143.93
tokens on average, while SFT and EPIC generate 111.25 and 123.48 tokens,
respectively.

\begin{table}[t]
\centering
\caption{
\textbf{Training overhead and response length.}
Compute measurements use batch size 32 on a single NVIDIA L40 GPU.
TFLOPs and time are reported per sample; peak memory is measured over the
full batch.
}
\label{tab:compute-cost}
\setlength{\tabcolsep}{5pt}
\begin{tabular}{lccc}
\toprule
Method & TFLOPs / Sample & Time / Sample & Peak Memory \\
\midrule
\textsc{SCOPD}  & 54.25 & 17.21\,s & 25.02\,GiB \\
\textsc{SCOPD+} & 65.84 & 17.54\,s & 25.13\,GiB \\
\bottomrule
\end{tabular}

\vspace{0.6em}

\begin{tabular}{lc}
\toprule
Method & Mean Output Tokens \\
\midrule
Vanilla (No Prune) & 155.27 \\
VisionZip 10\%     & 143.93 \\
SFT                 & 111.25 \\
EPIC                & 123.48 \\
\textsc{SCOPD}      & 153.00 \\
\textsc{SCOPD+}     & 155.70 \\
\bottomrule
\end{tabular}
\end{table}


\end{document}